\documentclass{article}
\usepackage{arxiv}
\usepackage[frozencache=true,cachedir=.]{minted}
\usepackage{dcolumn}
\usepackage[acronym]{glossaries}
\usepackage{graphicx}
\usepackage{hyperref}
\usepackage{imakeidx}  
\usepackage{latexsym}
\usepackage{mfirstuc}
\usepackage{multicol}
\usepackage{multirow}
\usepackage{rotating}
\usepackage{colortbl}
\usepackage{tabularx,booktabs}
\usepackage{url}
\usepackage{fancyvrb}
\usepackage[usenames]{xcolor}

\usepackage[utf8]{inputenc}
\usepackage{outlines}
\usepackage{wasysym}
\usepackage{etoolbox}
\AtBeginDocument{%
  \providecommand\BibTeX{{%
    \normalfont B\kern-0.5em{\scshape i\kern-0.25em b}\kern-0.8em\TeX}}}

\newminted[code]{python}{linenos, breaklines, fontsize=\footnotesize, stripnl=true}
\newif\ifdraft
\newcommand{\codeline}[1]{\mintinline{python}{#1}}

\newcommand{\svmperf}{SVM$^{\mathrm{perf}}$}
\newcommand{\ehdy}{E(HDy)$_{\mathrm{DS}}$}

\newcommand{\pdag}{\phantom{^{\dag}}}

\newcommand{\side}[1]{\begin{sideways}{#1}\end{sideways}}

\begin{document}

%% The "title" command has an optional parameter, allowing the author
%% to define a "short title" to be used in page headers.

\title{QuaPy: A Python-Based Framework for Quantification}

%% the authors and their affiliations.  Of note is the shared
%% affiliation of the first two authors, and the "authornote" and
%% "authornotemark" commands used to denote shared contribution to the
%% research.
\author{ 
\href{https://orcid.org/0002-0377-1025}{\includegraphics[scale=0.06]{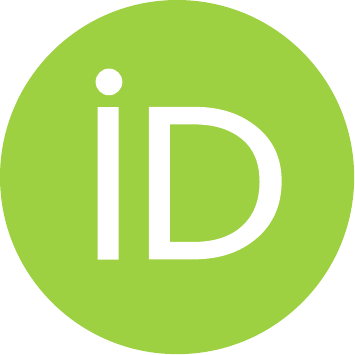}\hspace{1mm}Alejandro Moreo\thanks{Corresponding author}}, \hspace{3mm}
\href{https://orcid.org/0002-5725-4322}{\includegraphics[scale=0.06]{orcid.pdf}\hspace{1mm}Andrea Esuli}, \hspace{3mm}
\href{https://orcid.org/0003-4221-6427}{\includegraphics[scale=0.06]{orcid.pdf}\hspace{1mm}Fabrizio Sebastiani}\\
	Istituto di Scienza e Tecnologie dell'Informazione\\
	Consiglio Nazionale delle Ricerche\\
	Via Giuseppe Moruzzi 1 \\ Pisa, Italy, 56124\\
	\texttt{\{first.last\}@isti.cnr.it} \\
	%% examples of more authors
	\\
}

\maketitle

\begin{abstract}
  QuaPy is an open-source framework for performing
  \textit{quantification} (a.k.a.\ \textit{supervised prevalence
  estimation}), written in Python.  Quantification is the task of
  training \textit{quantifiers} via supervised learning, where a
  quantifier is a predictor that estimates the \textit{relative
  frequencies} (a.k.a.\ \textit{prevalence values}) of the classes of
  interest in a sample of unlabelled data. While quantification can be
  trivially performed by applying a standard classifier to each
  unlabelled data item and counting how many data items have been
  assigned to each class, it has been shown that this ``classify and
  count'' method is outperformed by methods specifically designed for
  quantification.  QuaPy
  % is based on the concept of \textit{data sample}, and
  provides implementations of
  % the most important concepts in the quantification literature,
  % including
  a number of baseline methods and advanced quantification methods, of
  routines for quantification-oriented model selection, of several
  broadly accepted evaluation measures, and of robust evaluation
  protocols routinely used in the field. QuaPy also makes available
  datasets commonly used for testing quantifiers, and offers
  visualization tools for facilitating the analysis and interpretation
  of the results.  The software is open-source and publicly available
  under a BSD-3 licence via
  GitHub\footnote{\url{https://github.com/HLT-ISTI/QuaPy}}, and can be
  installed via
  \texttt{pip}\footnote{\url{https://pypi.org/project/QuaPy/}}.
\end{abstract}

\keywords{Quantification, Supervised Prevalence Estimation, Learning
to Quantify, Supervised Learning, Python, Open Source}

% --------------------------------------------------------------------

\section{Introduction}

\noindent \textit{Quantification} (variously called \textit{learning
to quantify}, or \textit{supervised prevalence estimation}, or
\textit{class prior estimation}) is the task of training models
(``quantifiers'') that estimate the \textit{relative frequencies}
(a.k.a.\ \textit{prevalence values}) of the classes of interest in a
sample of unlabelled data items~\cite{Gonzalez:2017it}. For instance,
in a sample of 100,000 unlabelled tweets known to express opinions
about Donald Trump, such a model may be tasked to estimate the
percentage of these 100,000 tweets which display a \textsf{Positive}
stance towards Trump (and to do the same for classes \textsf{Neutral}
and \textsf{Negative}). In other words, quantification stands to
classification as aggregate data stand to individual
data. Quantification is of special interest in fields such as the
social sciences~\cite{Hopkins:2010fk},
epidemiology~\cite{King:2008fk}, market
research~\cite{Esuli:2010kx}, and ecological
modelling~\cite{Beijbom:2015yg}, since these fields are inherently
concerned with aggregate data; however, quantification is also useful
in applications outside these fields, such as in enforcing the
fairness of classifiers~\cite{Biswas:2019vn}, performing word sense
disambiguation~\cite{Chan2006}, allocating
resources~\cite{Forman:2008kx}, and improving the accuracy of
classifiers~\cite{Saerens:2002uq}.

Quantification can trivially be solved via classification, i.e., by
training a classifier, applying it to the unlabelled data items, and
counting how many data items have been assigned to each
class. However, there is by now abundant
evidence~\cite{Gonzalez:2017it} that this ``classify and count''
method delivers suboptimal quantification accuracy, and especially so
in scenarios characterized by \textit{distribution shift}, i.e., by
the fact that the class prevalence values in the training set are
substantially different from those in the set of unlabelled data. As a
result, quantification is no more considered just a by-product of
classification, and has evolved as a task in its own right; as such,
quantification has its own learning methods, model selection
protocols, evaluation measures, and evaluation protocols.

% estimating the \textit{relative frequency}
% (a.k.a. \textit{prevalence}, or \textit{prior probability}) of a set
% of predefined classes of interest given a target population of
% individuals.  The analysis focuses on data at the \textit{aggregate}
% level and, in a way, even disregards the treatment of
% \textit{individuals} during the analysis.  This forms the basis for
% the observation that resolving such a task as a mere by-product of
% classification leads to sub-optimal solutions, something which has
% conclusively been proved in related literature and which has given
% rise to the definition of \textit{quantification}
% (a.k.a. \textit{supervised prevalence estimation}, or
% \textit{learning to quantify}) as a task on its own right
% \cite{Gonzalez:2017it}. As such, quantification is investigated with
% specific methodologies, evaluation measures, and dedicated
% evaluation protocols.
In this paper we present QuaPy, a framework written in Python that
provides implementations of the most important tools for research,
development, and experimentation, in quantification.
%
% QuaPy roots on the concept of data sample, and provides
% implementations of most important concepts in quantification
% literature, such as the most important quantification baselines,
% many advanced quantification methods, quantification-oriented model
% selection, many evaluation measures and protocols used for
% evaluating quantification methods. QuaPy also integrates commonly
% used datasets and offers visualization tools for facilitating the
% analysis and interpretation of results.
%
% \subsection{A quick example:}
The following script can serve as a minimal working example of how
QuaPy is used. The script fetches a dataset of tweets, trains a
quantifier via the \textit{Adjusted Classify and Count} method (that
is meant to improve the prevalence estimates returned by a standard
classifier, here trained via logistic regression), and then evaluates
the quantifier in terms of the \textit{Absolute Error} (AE) between
the estimated and the true class prevalence values of the test set.

\begin{code}
  import quapy as qp
  from sklearn.linear_model import LogisticRegression

  data = qp.datasets.fetch_twitter('semeval16')

  # create an "Adjusted Classify & Count" quantifier
  model = qp.method.aggregative.ACC(LogisticRegression())
  model.fit(data.training)

  estim_prevalence = model.quantify(data.test.instances)
  true_prevalence  = data.test.prevalence()

  error = qp.error.ae(true_prevalence, estim_prevalence)
  print('Absolute Error (AE)', error)
\end{code}

\noindent As mentioned above, quantification is particularly useful in
scenarios where distribution shift may occur.
% After all, we would not be interested in estimating the class
% prevalence values of the test set if we could assume the IID
% assumption to hold strictly, since the sought prevalence would
% trivially coincide with the class prevalence of the training set.
Any quantification model should thus be tested across different data
samples characterized by different class prevalence values. QuaPy
implements sampling procedures and evaluation protocols that automate
this endeavour.
%
% \subsection{Features}

The paper is structured as follows. In Section~\ref{sec:methods} we
briefly describe the quantifier training methods included in QuaPy,
while in Section~\ref{sec:datasets} we present a number of datasets
that have been previously used in quantification research and that we
include in the QuaPy suite. Section~\ref{sec:eval} is devoted to
quantifier evaluation, and discusses the evaluation measures and
evaluation protocols that we make available within
QuaPy. Section~\ref{sec:modsel} turns to model selection, discussing
the hyperparameter optimization protocols implemented within QuaPy,
while Section~\ref{sec:plots} illustrates the tools that we make
available for visualizing the results of quantification
experiments. Section~\ref{sec:experiments} discusses some experiments
that we have carried out in order to showcase some among the features
of QuaPy.  In Section~\ref{sec:conclusion} we give some concluding
remarks.
%
% Some features of QuaPy include:
% \begin{itemize}
% \item Implementation of many popular quantification methods
%   (Section~\ref{sec:methods}),%(Classify-\&-Count variants, Expectation-Maximization, SVM-based variants for quantification, HDy, QuaNet, and Ensembles).
%   with native support for both binary and multi-class quantification
%   scenarios (Section~\ref{sec:single});
% \item Popular datasets (both textual and non-textual) used in the
%   quantification literature
%   (Section~\ref{sec:datasets});%, including:
%   %   \begin{itemize}
%   %  \item 32 UCI Machine Learning datasets.
%   %  \item 11 Twitter Sentiment datasets.
%   %  \item 3 Reviews Sentiment datasets.
%   %\end{itemize}
% \item Implementation of most commonly used quantifier evaluation
%   metrics
%   (Section~\ref{sec:errors}); %(e.g., MAE, MRAE, MSE, NKLD, etc.).
% \item Versatile functionality for performing evaluation based on
%   standard quantifier evaluation protocols
%   (Section~\ref{sec:protocols});
% \item Model selection functionality using quantification-oriented
%   loss measures (Section~\ref{sec:modsel});
% \item Visualization tools for analysing the results
%   (Section~\ref{sec:plots}).
% \end{itemize}
%

% --------------------------------------------------------------------

\section{Methods}
\label{sec:methods}

\noindent A quantifier is defined in QuaPy as a model that can be
\texttt{fit} on some training data, so that the fitted model can
estimate class prevalence values for unlabelled data.  More
specifically, a quantifier in QuaPy must inherit from the class
\codeline{BaseQuantifier}, and implement the following abstract
methods:
\begin{code}
  @abstractmethod
  def fit(self, data: LabelledCollection): ...

  @abstractmethod
  def quantify(self, instances): ...

  @abstractmethod
  def set_params(self, **parameters): ...

  @abstractmethod
  def get_params(self, deep=True): ...
\end{code}
\noindent The meaning of these functions should be familiar to anybody
accustomed to the \texttt{scikit-learn}
environment~\cite{scikit-learn}, since the class structure of QuaPy is
directly inspired by \texttt{scikit-learn}'s
``estimators''.\footnote{QuaPy's quantifiers do not inherit from
\texttt{scikit-learn}'s estimators due to one key difference that
makes the two incompatible. While a \texttt{scikit-learn} estimator's
\texttt{predict} method is expected to produce an array of $c$
predictions (with $c$ the number of classes) for each of the $n$ data
items in the input, the quantifier's \texttt{quantify} method is
instead requested to output one single vector of $c$ prevalence values
for a given sample of data items.} Functions \texttt{fit} and
\texttt{quantify} are used to train the model and to return class
prevalence
estimates, %(the reason why \texttt{scikit-learn}' structure has not been adopted as is in QuaPy responds to the fact that \texttt{scikit-learn}'s predict function is expected to return one output for each input element --e.g., a predicted label for each instance in a sample-- while in quantification the output for a sample is one single array of class prevalence values),
respectively, while functions \texttt{set\_params} and
\texttt{get\_params} allow a model-selecting routine (see
Section~\ref{sec:modsel}) to automate the process of hyperparameter
optimization.

Quantification methods can be classified as belonging to the
\textit{}{aggregative}, \textit{non-aggregative}, or \textit{meta}
classes. Aggregative methods are characterized by the fact that
quantification is obtained as an aggregation of the outputs returned
by a classification process for the individual documents.
Non-aggregative methods analyse instead the sample of unlabelled
documents as a whole, without resorting to the classification of
individual data items.  Finally, meta-quantifiers are built on top of
other quantifiers, and generate their predictions by analysing the
predictions made by the underlying quantifiers. We will briefly
present these three classes in the next three subsections.

% --------------------------------------------------------------------

\subsection{Aggregative methods}

\noindent Most of the methods proposed in the literature and included
in QuaPy are aggregative.  QuaPy models aggregative quantifiers by
means of the abstract class \texttt{AggregativeQuantifier}. This class
extends \texttt{BaseQuantifier}, providing a default implementation of
the \texttt{quantify} method based on the \texttt{aggregate} function,
that has to be implemented, i.e.,

\begin{code}
  def quantify(self, instances):
      classif_predictions = self.classify(instances)
      return self.aggregate(classif_predictions)

  @abstractmethod
  def aggregate(self, classif_predictions:np.ndarray): ...
\end{code}

\noindent Implementing an aggregative method only requires overriding
the \texttt{aggregate} method.  The \texttt{AggregativeQuantifier}
class implements the rest of the process, and is designed to work with
any \texttt{scikit-learn} estimator.  Working with packages or machine
learning tools other than \texttt{scikit-learn} only requires
overriding the \texttt{classify} method, which takes as input the
individual data items in the sample and returns the corresponding
classification predictions (see Section~\ref{sec:svmperf}).

\textit{Probabilistic} aggregative methods are a subclass of
aggregative methods, which, instead of the ``crisp'' decisions
returned by a categorical classifier, use the posterior probabilities
returned by a probabilistic classifier.  Probabilistic aggregative
methods inherit from the abstract class
\texttt{AggregativeProbabilisticQuantifier}, which extends
\texttt{AggregativeQuantifier}, by providing a default implementation
of the \texttt{quantify} method as follows:

\begin{code}
  def quantify(self, instances):
      classif_posteriors = self.posterior_probabilities(instances)
      return self.aggregate(classif_posteriors)
\end{code}

\noindent The method \texttt{posterior\_probabilities}, similarly to
the more general case, is designed to work together with the
\texttt{predict\_proba} method of any probabilistic classifier in
\texttt{scikit-learn}.  QuaPy also allows using the
\texttt{scikit-learn}'s crisp estimators that do not come with an
implementation of the \texttt{predict\_proba} method (e.g.,
\texttt{LinearSVC}).  In this case, the estimator is converted into a
probabilistic classifier by means of a \emph{calibration}
method~\cite{Platt99}.\footnote{In QuaPy this is automatically done by
wrapping the estimator in the \texttt{CalibratedClassifierCV} class.}
Packages other than \texttt{scikit-learn} can be used as well by
providing a custom implementation of the
\texttt{posterior\_probabilities} method (see
Section~\ref{sec:svmperf}).

One advantage of aggregative methods (probabilistic or not) is that
the evaluation according to any sampling procedure (e.g., the
artificial prevalence protocol -- see Section~\ref{sec:eval}) can be
carried out very efficiently, since the entire set of unlabelled items
can be pre-classified once for all at the beginning, and the
estimation of class prevalence values for different samples can
directly reuse these predictions, with no need to reclassify each
individual data item every time.  QuaPy takes advantage of this
property to drastically speed up any routine that has to do with
quantification on multiple samples drawn from the same set, as is
customarily the case in quantification, both in the performance
evaluation phase (Section~\ref{sec:eval}) and in the model selection
phase (Section~\ref{sec:modsel}).

% --------------------------------------------------------------------

\subsubsection{Classify \& Count and its variants}

\noindent QuaPy provides implementations for \textit{Classify \&
Count} (CC) and its variants, i.e.,

\begin{itemize}
\item CC (Classify \& Count), the simplest aggregative quantifier,
  that simply relies on the label predictions of a classifier to
  deliver class prevalence estimates;
\item ACC (Adjusted Classify \& Count)~\cite{Forman:2008kx}, the
  ``adjusted'' variant of CC, that corrects the predictions of CC
  according to the ``misclassification rates'' (see below) of the
  classifier;
\item PCC (Probabilistic Classify \& Count)~\cite{Bella:2010kx}, the
  probabilistic variant of CC that relies on the posterior
  probabilities returned by a probabilistic classifier;
\item PACC (Probabilistic Adjusted Classify \&
  Count)~\cite{Bella:2010kx}, which stands to PCC as ACC stands to
  CC.
\end{itemize}

% The following code serves as a complete example using CC equipped
% with a SVM as the classifier:

% \begin{code}
%import quapy as qp
%import quapy.functional as F
%from sklearn.svm import LinearSVC
%
%dataset = qp.datasets.fetch_twitter('hcr', pickle=True)
%training = dataset.training
% test = dataset.test

% # instantiate a classifier learner, in this case a SVM
% svm = LinearSVC()

% # instantiate a Classify & Count with the SVM
% # (an alias is available in qp.method.aggregative.ClassifyAndCount)
% model = qp.method.aggregative.CC(svm)
% model.fit(training)
% estim_prevalence = model.quantify(test.instances)
% \end{code}
% The same code could be used to instantiate an ACC, by simply
% replacing the instantiation of the model with:
% \begin{code}
% model = qp.method.aggregative.ACC(svm)
% \end{code}

\noindent Note that the adjusted variants (ACC and PACC) need to
estimate the parameters (the ``misclassification rates'') required for
performing the adjustment; the estimation
% (e.g., the \textit{true positive rate} and the \textit{false
% positive rate} in cases of binary classification)
uses a validation set carved out of the labelled set.  The specific
form of parameter optimization can be set at construction time or at
fitting time using the argument \texttt{val\_split}, either by
indicating a \texttt{float} in (0,1) specifying the fraction of the
training data to be used as a held-out validation set, or by
indicating an \texttt{int} specifying the number of folds to be used
in a $k$-fold cross-validation ($k$-FCV) process, or by explicitly
passing a set of instances to be used as the validation set (i.e., an
instance of \texttt{LabelledCollection} -- see
Section~\ref{sec:datasets}).

\subsubsection{Forman's variants of ACC}

\noindent QuaPy also provides implementations of a series of binary
quantification methods, proposed by Forman
in~\cite{Forman:2006uf,Forman:2008kx} as variations of ACC, and whose
goal is to bring improved stability to the denominator of the
adjustment.\footnote{In the binary case, the ACC adjustment comes down
to computing
$\hat{p}^{\mathrm{ACC}}(y)=\frac{\hat{p}^{\mathrm{CC}}(y)-\hat{\mathrm{fpr}}(y)}{\hat{\mathrm{tpr}}(y)-\hat{\mathrm{fpr}}(y)}$
in which $\hat{p}^{\mathrm{CC}}(y)$ is the prevalence of class $y$ as
estimated by CC, and $\hat{\mathrm{tpr}}(y)$ and
$\hat{\mathrm{fpr}}(y)$ stand for the \textit{true positive rate} and
\textit{false positive rate} of the classifier, as estimated in the
validation phase. The above-mentioned numerical instability arises
when $\hat{\mathrm{tpr}}(y)\approx \hat{\mathrm{fpr}}(y)$.}  The
methods are based on different heuristics for choosing a decision
threshold that would allow for more true positives and many more false
positives, on the grounds this would deliver larger denominators.
% , on the grounds that such a system would perform better in
% quantification (though at the cost of dropping classification
% performance).

QuaPy implements the methods X (which looks for the threshold that
yields $\mathrm{tpr}(y)=1-\mathrm{fpr}(y)$), MAX (which looks for the
threshold that maximizes $\mathrm{tpr}(y)-\mathrm{fpr}(y)$), T50
(which looks for the threshold that makes $\mathrm{tpr}(y)$ closest to
0.5). QuaPy also implements MS (Median Sweep), a method that generates
class prevalence estimates for all decision thresholds and returns the
median of them all; and MS2, a variant that computes the median only
for cases in which $\mathrm{tpr}(y)-\mathrm{fpr}(y)>0.25$.

% --------------------------------------------------------------------

\subsubsection{The Saerens-Latinne-Decaestecker algorithm}

\noindent The Saerens-Latinne-Decaestecker (SLD)
algorithm~\cite{Saerens:2002uq,Esuli:2020le} (sometimes also
called EMQ, for \textit{Expectation Maximization Quantifier}) is a
probabilistic quantifier-generating method.  SLD consists of using the
well-known Expectation Maximization algorithm to iteratively update
the posterior probabilities generated by a probabilistic classifier
and the class prevalence estimates obtained via maximum-likelihood
estimation, in a mutually recursive way, until convergence.
% shift the estimation of class prevalence values as observed in the
% training set, to one that maximizes the likelihood on the test data.
Although this method was originally proposed for improving the quality
of the posterior probabilities returned by a probabilistic classifier,
and not for improving its class prevalence estimates, SLD has proven
to be among the most effective quantifiers in many
experiments~\cite{moreo2021re,Moreo:2020mk,Schumacher}.

% --------------------------------------------------------------------

\subsubsection{The HDy method}

\noindent HDy~\cite{Gonzalez-Castro:2013fk} is a probabilistic method
for training binary quantifiers, that models quantification as the
problem of minimizing the divergence (in terms of the
\textit{Hellinger Distance}) between two cumulative distributions of
posterior probabilities returned by the classifier. One of the
distributions is generated from the unlabelled examples and the other
is generated from a validation set. This latter distribution is
defined as a mixture of the class-conditional distributions of the
posterior probabilities returned for the positive and negative
validation examples, respectively. The parameters of the mixture thus
represent the estimates of the class prevalence values.

Since the method requires a validation set to estimate the parameters
of the mixture model, the constructor and \texttt{fit} method of HDy
receive as input the argument \texttt{val\_split}, whose semantics is
the same as in ACC and PACC.

% \todo{short description}

% HDy requires a validation set to estimate parameter for the mixture
% model. %Just like ACC and PACC, this quantifier receives a val_split argument in the constructor (or in the fit method, in which case the previous value is overridden) that can either be a float indicating the proportion of training data to be taken as the validation set (in a random stratified split), or a validation set (i.e., an instance of \texttt{LabelledCollection}) itself.

% Differently from all \todo{check if it is all, or most} other method
% implemented in QuaPy, which work for both binary and multi-class
% cases, HDy was proposed as a binary quantifier and the
% implementation accepts only binary datasets.

% The following code shows an example of use:

% \begin{code}
% import quapy as qp
% from sklearn.linear_model import LogisticRegression

% # load a binary dataset
% dataset = qp.datasets.fetch_reviews('hp', pickle=True)
% qp.data.preprocessing.text2tfidf(dataset, min_df=5, inplace=True)

% model = qp.method.aggregative.HDy(LogisticRegression())
% model.fit(dataset.training)
% estim_prevalence = model.quantify(dataset.test.instances)
% \end{code}

% --------------------------------------------------------------------

\noindent

% --------------------------------------------------------------------

\subsubsection{Quantifiers based on Explicit Loss Minimization}
\label{sec:svmperf}

\noindent The quantifiers based on \textit{Explicit Loss Minimization}
(ELM) represent a family of methods based on structured output
learning; these quantifiers rely on classifiers that have been
optimized using a quantification-oriented loss measure. QuaPy
implements the following ELM-based methods, all relying on Joachims'
\svmperf\ structured output learning
algorithm~\cite{Joachims:2006rw}:\footnote{QuaPy includes the tools to
automatically patch the original \svmperf\ code in order to add the
quantification-oriented loss functions.}

\begin{itemize}
\item SVM(Q), which attempts to minimize the $Q$ loss, that combines a
  classification-oriented loss and a quantification-oriented loss, as
  proposed in~\cite{Barranquero:2015fr};
\item SVM(KLD), which attempts to minimize the Kullback-Leibler
  Divergence, as proposed in~\cite{Esuli:2010fk} and as first used
  in~\cite{Esuli:2015gh};
\item SVM(NKLD), which attempts to minimize a version of the
  Kullback-Leibler Divergence normalized via the logistic function, as
  first used in~\cite{Esuli:2015gh};
\item SVM(AE), which uses Absolute Error as the loss, as first used
  in~\cite{moreo2021re};
\item SVM(RAE), which uses Relative Absolute Error as the loss, as
  first used in~\cite{moreo2021re}.

\end{itemize}

% \todo{Che vuol dire questa frase qui sotto?}  \alexcomment{Pensavo
% che SVMAE e SVMRAE non fossero state usate in passato...}  the last
% two methods (SVMAE and SVMRAE) have been implemented in QuaPy in
% order to make available ELM variants for what nowadays are
% considered the most well-behaved evaluation metrics in
% quantification.

% In order to make these models work, you would need to run the script
% prepare_svmperf.sh (distributed along with QuaPy) that downloads
% SVMperf' source code, applies a patch that implements the
% quantification oriented losses, and compiles the sources.

% If you want to add any custom loss, you would need to modify the
% source code of SVMperf in order to implement it, and assign a valid
% loss code to it. Then you must re-compile the whole thing and
% instantiate the quantifier in QuaPy as follows:

% \begin{itemize}
% \item you can either set the path to your custom
%   \texttt{svm\_perf\_quantification} implementation
% \item in the environment variable, or as an argument to the
%   constructor of ELM \texttt{qp.environ['SVMPERF\_HOME'] =
%   './path/to/svm\_perf\_quantification'}

% \item assign an alias to your custom loss and the id you have
%   assigned to it \texttt{svmperf = qp.classification.svmperf.SVMperf
%   svmperf.valid\_losses['mycustomloss'] = 28}

% \item instantiate the ELM method indicating the loss \texttt{model =
%   qp.method.aggregative.ELM(loss='mycustomloss')}
% \end{itemize}

\noindent All ELM-based methods can train binary quantifiers only,
since they rely on \svmperf, which is an inherently binary
system. %(see Section~\ref{sec:single}).
However, QuaPy allows the conversion of binary quantifiers into
multi-class quantifiers (see Section~\ref{sec:single}).

% --------------------------------------------------------------------

\subsection{Methods for training meta-quantifiers}

\noindent Meta-quantifiers base their estimates on the estimates
produced by other quantifiers, and are defined in the
\texttt{qp.method.meta} module.
% By meta models we mean quantification methods that are defined on
% top of other quantification methods, and As such, they do not
% squarely belong to the aggregative nor the non-aggregative group
% (indeed, meta models could use quantifiers from any of those
% groups). Meta models are implemented in the \texttt{qp.method.meta
% }module.

% --------------------------------------------------------------------

\subsubsection{Ensembles:}

\noindent A quantification ensemble receives as input any
quantification method (any instance of \texttt{BaseQuantifier}).
QuaPy implements some among the ``ensemble'' variants proposed
in~\cite{Perez-Gallego:2017wt,Perez-Gallego:2019vl}, that train
different members of the ensemble using different samples of the
original training set; in particular:
\begin{itemize}
\item \texttt{Averaging} (\codeline{policy='ave'}, default): computes
  class prevalence estimates as the average of the estimates returned
  by the base quantifiers.
\item \texttt{Training Prevalence} (\codeline{policy='ptr'}): applies
  a dynamic selection to the ensemble's members by retaining only
  those members such that the class prevalence values in the samples
  they use as training set are closest to preliminary class prevalence
  estimates computed as the average of the estimates of all the
  members. The final estimate is recomputed by considering only the
  selected members.
\item \texttt{Distribution Similarity} (\codeline{policy='ds'}):
  performs a dynamic selection of base members %(as before), but
  by retaining the members trained on samples whose distribution of
  posterior probabilities is closest, in terms of the Hellinger
  Distance, to the distribution of posterior probabilities in the test
  sample;
  % training prevalence and the preliminary class prevalence estimate
  % discussed above; \fabsebcomment{E' la stessa della precedente,
  % solo che usa la Hellinger Distance al posto di un'altra
  % prevalence? quale prevalence?}
\item \texttt{Performance} (\codeline{policy='<any-error-metric>'}):
  performs a static selection of the ensemble members by retaining
  those that minimize a quantification error measure, which is passed
  as an argument.
\end{itemize}

% The following code shows how to instantiate an Ensemble of 30
% Adjusted Classify & Count (ACC) quantifiers operating with a
% Logistic Regressor (LR) as the base classifier, and using the
% average as the aggregation policy (see the original article for
% further details). The last parameter indicates to use all processors
% for parallelization.

% \begin{code}
% import quapy as qp
% from quapy.method.aggregative import ACC
% from quapy.method.meta import Ensemble
% from sklearn.linear_model import LogisticRegression

% dataset = qp.datasets.fetch_UCIDataset('haberman')

% model = Ensemble(quantifier=ACC(LogisticRegression()), size=30, policy='ave', n_jobs=-1)
% model.fit(dataset.training)
% estim_prevalence = model.quantify(dataset.test.instances)
% \end{code}

% Other aggregation policies implemented in QuaPy include:

% \begin{itemize}
% \item 'ptr' for applying a dynamic selection based on the training
%   prevalence of the ensemble's members
% \item 'ds' for applying a dynamic selection based on the Hellinger
%   Distance
% \item any valid quantification measure (e.g., 'mse') for performing
%   a static selection based on the performance estimated for each
%   member of the ensemble in terms of that evaluation metric.
% \end{itemize}

\noindent When using either dynamic or static selection policies, one
has to set the \texttt{red\_size} parameter, which defines the number
of members that have to be retained.

% Please, check the model selection wiki if you want to optimize the
% hyperparameters of ensemble for classification or quantification.

% --------------------------------------------------------------------

\subsubsection{The QuaNet recurrent quantifier:}

\noindent QuaPy provides an implementation of QuaNet, a
deep-learning-based method for performing quantification on samples of
textual documents, presented in~\cite{Esuli:2018rm}.\footnote{In order
to use QuaNet within QuaPy, the \texttt{torch} framework for deep
learning~\cite{NEURIPS2019_9015} has to be installed.} QuaNet
processes as input a list of document embeddings (see below), one for
each unlabelled document along with their posterior probabilities
generated by a probabilistic classifier. The list is processed by a
bidirectional LSTM that generates a sample embedding (i.e., a dense
representation of the entire sample), which is then concatenated with
a vector of class prevalence estimates produced by an ensemble of
simpler quantification methods (CC, ACC, PCC, PACC, SLD). This vector
is then transformed by a set of feed-forward layers, followed by ReLU
activations and dropout, to compute the final estimations.

QuaNet thus requires a probabilistic classifier that can provide
embedded representations of the inputs. QuaPy offers a basic
implementation of such a classifier, based on convolutional neural
networks, that returns its next-to-last representation as the document
embedding. The following is a working example showing how to index a
textual dataset (see Section~\ref{sec:datasets}) and how to
instantiate QuaNet:

\begin{code}
  import quapy as qp
  from quapy.method.meta import QuaNet
  from classification.neural import NeuralClassifierTrainer, CNNnet

  qp.environ['SAMPLE_SIZE'] = 500
  
  # load the kindle dataset as plain text, and convert words 
  # to numerical indexes
  dataset = qp.datasets.fetch_reviews('kindle', pickle=True)
  qp.data.preprocessing.index(dataset, min_df=5, inplace=True)

  cnn = CNNnet(dataset.vocabulary_size, dataset.n_classes)
  learner = NeuralClassifierTrainer(cnn, device='cuda')
  model = QuaNet(learner, sample_size=500, device='cuda')
\end{code}

% \begin{code}
% import quapy as qp
% from quapy.method.meta import QuaNet
% from classification.neural import NeuralClassifierTrainer, CNNnet

% # use samples of 100 elements
% qp.environ['SAMPLE_SIZE'] = 100

% # load the kindle dataset as text, and convert words to numerical indexes
% dataset = qp.datasets.fetch_reviews('kindle', pickle=True)
% qp.data.preprocessing.index(dataset, min_df=5, inplace=True)

% # the text classifier is a CNN trained by NeuralClassifierTrainer
% cnn = CNNnet(dataset.vocabulary_size, dataset.n_classes)
% learner = NeuralClassifierTrainer(cnn, device='cuda')

% # train QuaNet
% model = QuaNet(learner, qp.environ['SAMPLE_SIZE'], device='cuda')
% model.fit(dataset.training)
% estim_prevalence = model.quantify(dataset.test.instances)
% \end{code}

% --------------------------------------------------------------------

\subsection{Using binary quantifiers in multi-class quantification}
\label{sec:single}

\noindent QuaPy allows a set of binary quantifiers, one for each
class, to be assembled into a single-label multi-class quantifier, by
adopting a ``one-vs-all" strategy.  This takes the form of computing
prevalence estimates independently for each class (i.e., via binary
quantification) via independently trained binary quantifiers, and then
normalizing the resulting vector of prevalence values (via
L1-normalization) so that these values sum up to one.
% (examples of this may be found, e.g., in~\cite{Gao:2016uq}).
In QuaPy this is possible by wrapping any binary
quantifier %(the ELM method SVMQ in the following example)
within a \texttt{OneVsAll} object.  For example, a quantifier defined
as \codeline{model=OneVsAll(SVMQ())} will allow \texttt{SVMQ} to work
with single-label multiclass datasets.

% \begin{code}
% import quapy as qp
% from quapy.method.aggregative import SVMQ, OneVsAll

% # load a multi-class dataset (this one contains 3 classes)
% dataset = qp.datasets.fetch_twitter('hcr', pickle=True)

% # let qp know where svmperf is
% qp.environ['SVMPERF_HOME'] = '../svm_perf_quantification'

% model = OneVsAll(SVMQ(), n_jobs=-1)  # run them on parallel
% model.fit(dataset.training)
% estim_prevalence = model.quantify(dataset.test.instances)
% \end{code}

% --------------------------------------------------------------------

\section{Datasets}
\label{sec:datasets}

\noindent QuaPy makes available a number of datasets that have been
used for experimentation purposes in the quantification literature,
and specifically:\footnote{All these datasets have a corresponding
\texttt{fetch} method in QuaPy that automatically downloads the
dataset from a public repository, and caches it for reuse.}
\begin{itemize}
\item \textbf{Reviews}: a collection of 3 datasets of customer reviews
  about (1) Kindle devices (\textsc{Kindle}), (2) the Harry Potter's
  book series (\textsc{HP}), both already used in~\cite{Esuli:2015gh},
  and (3) the well-known IMDB movie reviews dataset
  (\textsc{IMDB})~\cite{Maas2011}. All reviews are classified
  according to (binary) sentiment polarity. The number of training
  documents range from 3821 (\textsc{Kindle}) to 25000 (\textsc{IMDB})
  and present examples in which labelled data are balanced
  (\textsc{IMDB}, 50\% positives), imbalanced (\textsc{Kindle}, 92\%
  positives), and severely imbalanced (\textsc{HP}, 98\% positives).
  
\item \textbf{Twitter Sentiment}: 11 datasets of tweets labelled by
  sentiment, as used in~\cite{Gao:2015ly}.  The raw text of the tweets
  is not available due to Twitter's Terms of Service, and tweets are
  instead provided as \emph{tf-idf}-weighted vectors.  Similarly to
  the Reviews datasets, these are high-dimensional datasets, with
  dimensionalities ranging from 199,151 to 1,215,742.
  % The SemEval datasets for year 2013 to 2015 share the same training
  % set, and have a different test set for each edition of the track.
  These datasets use three sentiment labels (\textsf{Positive},
  \textsf{Neutral}, \textsf{Negative}), and are thus useful for
  testing non-binary quantification methods.
  
\item \textbf{UCI}: 33 binary datasets from the UCI Machine Learning
  repository~\cite{Dua:2019}, as used
  in~\cite{Perez-Gallego:2017wt}.\footnote{Some of these datasets, in
  the original form as made available in the UCI Machine Learning
  repository, are not binary, but the authors
  of~\cite{Perez-Gallego:2017wt} have transformed each $n$-ary dataset
  into $n$ binary datasets according to a ``one-vs-all'' policy; the
  datasets we make available are the binary ones as generated by the
  authors of~\cite{Perez-Gallego:2017wt}.}  Differently from the
  previous datasets, these non-textual datasets are low-dimensional
  (with dimensionalities ranging from 3 to 256), thus providing
  diversity, in terms of type of data, with respect to to the previous
  two sets of datasets.
\end{itemize}
\noindent QuaPy defines a simple \texttt{Dataset} interface that
allows importing any custom dataset into the QuaPy environment.  A
\texttt{Dataset} object in QuaPy is essentially a pair of
\texttt{LabelledCollection} objects, playing the role of the training
set and of the test set, respectively. \texttt{LabelledCollection} is
a data class consisting of the
% \aesucomment{(iterable)} \fabsebcomment{??????} \aesucomment{capisco
% il tuo dubbio, secondo me questo (iterable) si può levare. È molto
% più utile e importante la frase successiva sul fatto che implementa
% il sampling (l'idea è che se uno ha un dataset questa class ci
% realizza il sampling sopra fatto bene e gratis).}
instances and labels.  This class implements the core sampling
functionality in QuaPy, which is then exploited by the evaluation
tools (Section~\ref{sec:protocols}) and by the model selection tools
(Section~\ref{sec:modsel}).

From a \texttt{LabelledCollection}, QuaPy allows to easily produce new
samples at desired class prevalence values, i.e.,

\begin{code}
  sample_size = 100
  prev = [0.4, 0.1, 0.5]  # prevalence values for 3 classes
  sample = data.sampling(sample_size, *prev)
\end{code}

\noindent QuaPy supports the definition of samples consistent across
runs, in order to allow testing different quantification methods on
the very same
samples.% by sampling and retaining the indexes, that can then be used to generate the sample using the function \texttt{data.sampling\_index} to generate indexes, and then \texttt{sample.sampling\_from\_index} to effectively produce the sample.

% QuaPy also implements the artificial sampling protocol that produces
% (via a Python's generator) a series of LabelledCollection objects
% with equidistant prevalence values ranging across the entire
% prevalence spectrum in the simplex space:

% \begin{code}
% for sample in data.artificial_sampling_generator(sample_size=100, n_prevalence values=5):
%   print(F.strprev(sample.prevalence(), prec=2))
% \end{code}  
  
% produces one sampling for each (valid) combination of prevalence
% values originating from splitting the range [0,1] into n_prevalence
% values=5 points (i.e., [0, 0.25, 0.5, 0.75, 1]), that is:

% \begin{verbatim}
% [0.00, 0.00, 1.00]
% [0.00, 0.25, 0.75]
% [0.00, 0.50, 0.50]
% [0.00, 0.75, 0.25]
% [0.00, 1.00, 0.00]
% [0.25, 0.00, 0.75]
% ...
% [1.00, 0.00, 0.00]

% \end{verbatim}

% --------------------------------------------------------------------

\section{Evaluation}
\label{sec:eval}

% Quantification is an appealing tool in scenarios of dataset shift,
% and particularly in scenarios of prior-probability shift.  That is,
% the interest in estimating the class prevalence values arises under
% the belief that those class prevalence values might have changed
% with respect to the ones observed during training.  In other words,
% one could simply return the training prevalence as a predictor of
% the test prevalence if this change is assumed to be unlikely (as is
% the case in general scenarios of machine learning governed by the
% iid assumption).

\noindent Evaluating a quantifier requires measuring how good it is at
predicting the class prevalence values of a test sample, which may
have different class prevalence values than those observed on the
training data.

The evaluation of quantifiers is a complex task, since it depends on
many aspects.
% that may be more or less relevant with respect to the specific
% application one is facing.

For example, the same difference, in absolute value, between the true
and the predicted prevalence values may have a different ``cost''
depending on the original true prevalence value: predicting 0.5
prevalence when the true prevalence is 0.49 can be considered, in some
application contexts, a less blatant error than predicting a
prevalence of 0.01 when the true prevalence is 0.00. In some other
application contexts, though, the two above-mentioned estimation
errors may be considered equally
serious~\cite{Sebastiani:2020qf}. This means that sometimes we may
want to use a certain evaluation measure and some other times we may
want to use a different one.

Additionally, for some application contexts we may be interested in
measuring the quantification error only on samples whose class
prevalence values do not differ too much from those of the training
set, because we assume distribution shift, in practice, to always be
limited in magnitude. Conversely, in some other application contexts,
we may want to test our quantifiers also in situations characterized
by extreme values of distribution shift, because we expect our
environment to be characterized by high variability, and because we
want our quantifiers to be robust also to possibly extreme amounts of
shift.

As a result, an environment for experimenting with quantification must
not only be endowed with several evaluation measures, but it also must
allow the experimentation to be carried out according to different
evaluation protocols.

% --------------------------------------------------------------------

\subsection{Error measures}
\label{sec:errors}

\noindent Several error measures have been proposed in the
literature~\cite{Sebastiani:2020qf}, and QuaPy implements a rich set
of them:
\begin{itemize}
\item \texttt{ae}: absolute error
\item \texttt{rae}: relative absolute error
\item \texttt{se}: squared error
\item \texttt{kld}: Kullback-Leibler Divergence
\item \texttt{nkld}: normalized Kullback-Leibler Divergence
\end{itemize}

\noindent Functions \texttt{mae}, \texttt{mrae}, \texttt{mse},
\texttt{mkld}, and \texttt{mnkld} are also available, which return the
average values of the same measures across different samples.  For
aggregative quantifiers, also the $F_1$ and ``vanilla accuracy''
measures are available for measuring the quality of the underlying
classifiers.
% Some error measures for classification are also available:

% \begin{itemize}
% \item acce: accuracy error ($1-$accuracy)
% \item f1e: $F_1$ score error ($1-F_1$)
% \end{itemize}

% The error functions implement the following interface, e.g.:

% mae(true\_prevs, prevs\_hat)

% in which the first argument is a ndarray containing the true
% prevalence values, and the second argument is another ndarray with
% the estimations produced by some method.

Some error functions, e.g., \texttt{rae}, \texttt{kld}, and
\texttt{nkld}, and their averaged versions, are undefined for extreme
prevalence values (i.e., 0 and 1), and are numerically unstable for
prevalence values close to these extremes.  A common solution to this
problem is to perform smoothing, i.e., adding to each (true or
predicted) prevalence value a small amount, and then normalizing.  A
traditional smoothing value from the literature is $1/2T$, where $T$
is the size of the sample.  QuaPy supports setting the smoothing value
as an environment variable (\codeline{qp.environ['SAMPLE_SIZE']}), or
passing it as an argument of the error measure.
% In those cases, there is a third argument, e.g.:

% mrae(true\_prevs, prevs\_hat, eps=None)

% indicating the value for the smoothing parameter
% epsilon. Traditionally, this value is set to $1/(2T)$ in past
% literature, with $T$ the sampling size. One could either pass this
% value to the function each time, or to set a QuaPy's environment
% variable \texttt{SAMPLE\_SIZE} once, and omit this argument
% thereafter (recommended); e.g.:

% \begin{code}
% qp.environ['SAMPLE_SIZE'] = 100  # once for all
% true_prev = np.asarray([0.5, 0.3, 0.2])  # let's assume 3 classes
% estim_prev = np.asarray([0.1, 0.3, 0.6])
% error = qp.error.mrae(true_prev, estim_prev)
% print(f'mrae({true_prev}, {estim_prev}) = {error:.3f}')
% \end{code}

% will print:

% \begin{verbatim}
% mrae([0.500, 0.300, 0.200], [0.100, 0.300, 0.600]) = 0.914
% \end{verbatim}

% Finally, it is possible to instantiate QuaPy's quantification error
% functions from strings using, e.g.:

% \begin{code}
% error_function = qp.error.from_name('mse')
% error = error_function(true_prev, estim_prev)
% \end{code}

% --------------------------------------------------------------------

\subsection{Evaluation protocols}
\label{sec:protocols}

\noindent QuaPy implements both the \textit{natural prevalence
protocol} (NPP) and the \textit{artificial prevalence protocol} (APP).

In the NPP, the test set is sampled randomly, so that most samples
exhibit class prevalence values not to different from those of the
test set.
% \fabsebcomment{Dire che la distribuzione sulle distribuzioni è
% approssimata da una Gaussiana?}

In the APP, the test set is instead sampled in a controlled way, in
order to generate samples characterized by different, pre-specified
prevalence values, so as to cover, with uniform probability, the full
spectrum of class prevalence values.  In the APP the user specifies
the number of equidistant points to be generated from the interval
[0,1].  For example, if \texttt{n\_prevs=11} then, for each class, the
prevalence values [0.0, 0.1, ..., 0.9, 1.0] will be used. This means
that, for two classes, the number of different sampled prevalence
values will be 11 (since, once the prevalence of one class is
determined, the other one is also). For 3 classes, the number of valid
combinations can be obtained as 11 + 10 + ... + 1 = 66.  The number of
valid combinations (i.e., that sum up to one) that will be produced
for a given value of \texttt{n\_prevpoints} across \texttt{n\_classes}
can be determined by invoking
\texttt{quapy.functional.num\_prevalence\_combinations}, e.g.:

\begin{code}
  import quapy.functional as F
  n_prevs = 21  # [0, 0.05, 0.1, ..., 0.95, 1]
  n_classes = 4
  repeats = 1
  n = F.num_prevalence_combinations(n_prevs, n_classes, repeats)
\end{code}

\noindent In this example, $n=1771$.  The last argument,
\texttt{n\_repeats}, sets the number of samples that will be generated
for any valid combination (typical values are
% , e.g., 1 for a single sample, or
10 or higher, in order to support the computation of standard
deviations and to perform statistical significance tests).

One can instead work the other way around, i.e., set an evaluation
budged so as to obtain the number of prevalence values that will
generate a number of samples close but no higher than the fixed
budget, e.g.:

\begin{code}
  budget = 5000
  n_classes = 4
  repeats = 1
  n_prevs = F.get_nprevpoints_approximation(budget, n_classes, repeats)
  n = F.num_prevalence_combinations(n_prevs, n_classes, repeats)
\end{code}

% print(f'by setting n_prevpoints={n_prevpoints} the number of
% evaluations for {n_classes} classes will be {n}')

\noindent Here the function \texttt{get\_nprevpoints\_approximation}
determines that for the given budget and 4 classes, by setting
\texttt{n\_prevpoints}$=30$ the number of samples will be
\texttt{n}$=4960$.

QuaPy implements evaluation functions that allow the user to either
specify the \texttt{n\_prevpoints} value or an evaluation budget.  The
following script shows a full example in which a PACC model relying on
a classifier trained via logistic regression, is tested on the
\textsc{HP} dataset by means of the APP protocol on samples of size
500, setting a budget of 1000 test samples, in terms of various
evaluation metrics (\texttt{mae}, \texttt{mrae}, \texttt{mkld}).

\begin{code}
  import quapy as qp
  import quapy.functional as F
  from sklearn.linear_model import LogisticRegression

  # setting this environment variable allows some
  # error metrics (e.g., mrae) to be smoothed
  qp.environ["SAMPLE_SIZE"] = 500

  dataset = qp.datasets.fetch_reviews('hp', tfidf=True, min_df=5)

  training = dataset.training
  test = dataset.test

  lr = LogisticRegression()
  pacc = qp.method.aggregative.PACC(lr)

  pacc.fit(training)

  df = qp.evaluation.artificial_prevalence_report(
      pacc,  # the quantification method
      test,  # the test set on which the method will be evaluated
      sample_size=500,  # indicates the size of samples to be drawn
      eval_budget=1000,  # total number of samples to generate
      n_repetitions=10,  # number of samples for each prevalence
      n_jobs=-1,  # the number of parallel workers (-1 for all CPUs)
      random_seed=42,  # allows replicating test samples across runs
      error_metrics=['mae', 'mrae', 'mkld'],  # evaluation metrics
      verbose=True  # set to True to show some standard-line outputs
)
\end{code}

\noindent The resulting report is a \texttt{pandas} dataframe:
%
% import pandas as pd pd.set_option('display.expand_frame_repr',
% False) pd.set_option("precision", 3) df['estim-prev'] =
% df['estim-prev'].map(F.strprev) print(df)
%
\begin{small}
\begin{verbatim}
       true-prev   estim-prev    mae    mrae  mkld 
    0 [0.0, 1.0] [0.000, 1.000] 0.000  0.000 0.000 
    1 [0.0, 1.0] [0.000, 1.000] 0.000  0.000 0.000 
  ...  ...  ...
  ...  ...  ...
  998 [1.0, 0.0] [0.914, 0.086] 0.086 43.243 0.086
  999 [1.0, 0.0] [0.906, 0.094] 0.094 47.069 0.094
\end{verbatim}
\end{small}
%
% \noindent from which statistics can be computed.
%
% \begin{verbatim}
% true-prev    0.500
% mae          0.035
% mrae         2.578
% mkld         0.009
% dtype: float64
% \end{verbatim}
%
% Other evaluation functions include:
%
% \begin{itemize}
% \item \texttt{artificial\_sampling\_eval}: that computes the
%   evaluation for a given evaluation metric, returning the average
%   instead of a dataframe.
% \item \texttt{artificial\_sampling\_prediction}: that returns two
%   \texttt{np.arrays} containing the true prevalence values and the
%   estimated prevalence values.
% \end{itemize}
% See the documentation for further details.

% --------------------------------------------------------------------

\section{Model selection}
\label{sec:modsel}

% As a supervised machine learning task, quantification methods can
% strongly depend on a careful choice of model hyper-parameters.  The
% process whereby those hyper-parameters are chosen is typically known
% as Model Selection, and typically consists of testing different
% settings and picking the one that performed best in a held-out
% validation set in terms of any given evaluation measure.
%
% \subsection{Targeting a Quantification-oriented loss}
%
% The task being optimized determines the evaluation protocol, i.e.,
% the criteria according to which the performance of any given method
% for solving is to be assessed. As a task on its own right,
% quantification should impose its own model selection strategies,
% i.e., strategies aimed at finding appropriate configurations
% specifically designed for the task of quantification.
%
\noindent Quantification has long been regarded as a by-product of
classification, which means that the model selection (i.e.,
hyperparameter optimization) strategies customarily adopted in
quantification have simply been borrowed from classification.  It has
been argued in~\cite{moreo2021re} that specific model selection
strategies should be adopted for quantification. That is, model
selection strategies for quantification should minimize
quantification-oriented loss measures, and be carried out on a variety
of scenarios exhibiting different degrees of distribution shift.

QuaPy supports quantification-oriented model selection by
implementing, in the class \texttt{qp.model\_selection.GridSearchQ}, a
grid-search exploration over the space of hyperparameter combinations
that evaluates each such combination by means of a given
quantification-oriented error metric (see Section~\ref{sec:errors}),
and according to either the APP (the default value) or the NPP.

The following is an example of quantification-oriented model selection
using \texttt{GridSearchQ}.  In this example, model selection is
performed with a fixed budget of 1000 evaluations for each combination
of hyperparameters.  The loss function to miminize is MAE, a
quantification-oriented error measure, as evaluated on randomly drawn
samples at equidistant prevalence values covering the entire spectrum
(APP protocol) on a stratified held-out portion consisting of 40\% of
the training set.  \footnote{Classification-oriented model selection
can be done in QuaPy for aggregative quantifiers by simply using
\texttt{scikit-learn}'s \texttt{GridSearchCV} method on the base
Estimator.}

\begin{code}
  import quapy as qp
  from quapy.method.aggregative import PCC
  from sklearn.linear_model import LogisticRegression
  import numpy as np

  dataset = qp.datasets.fetch_reviews('hp', tfidf=True, min_df=5)

  # model selection with the APP
  model = qp.model_selection.GridSearchQ(
      model=PCC(LogisticRegression()),
      param_grid={'C': np.logspace(-4,5,10), 
                  'class_weight': ['balanced', None]},
      sample_size=500,
      protocol='app',  
      eval_budget=1000,
      error='mae',
      refit=True,  # retrain on the whole labelled set once done
      val_split=0.4,
  ).fit(dataset.training)

  # evaluation in terms of MAE
  results = qp.evaluation.artificial_prevalence_protocol(
      model,
      dataset.test,
      sample_size=500,
      n_prevpoints=101,
      n_repetitions=10,
      error_metric='mae'
  )

  print(f'best hyper-params={model.best_params_}')
  print(f'MAE={results:.5f}')
\end{code}

\noindent In this example, the system returns:
\begin{small}
\begin{verbatim}
  best hyper-params={'C': 0.1, 'class_weight': 'balanced'} 
  MAE=0.20342
\end{verbatim}
\end{small}

\section{Result visualization}
\label{sec:plots}

\begin{figure*}
  \centering \includegraphics[width=0.45\textwidth]{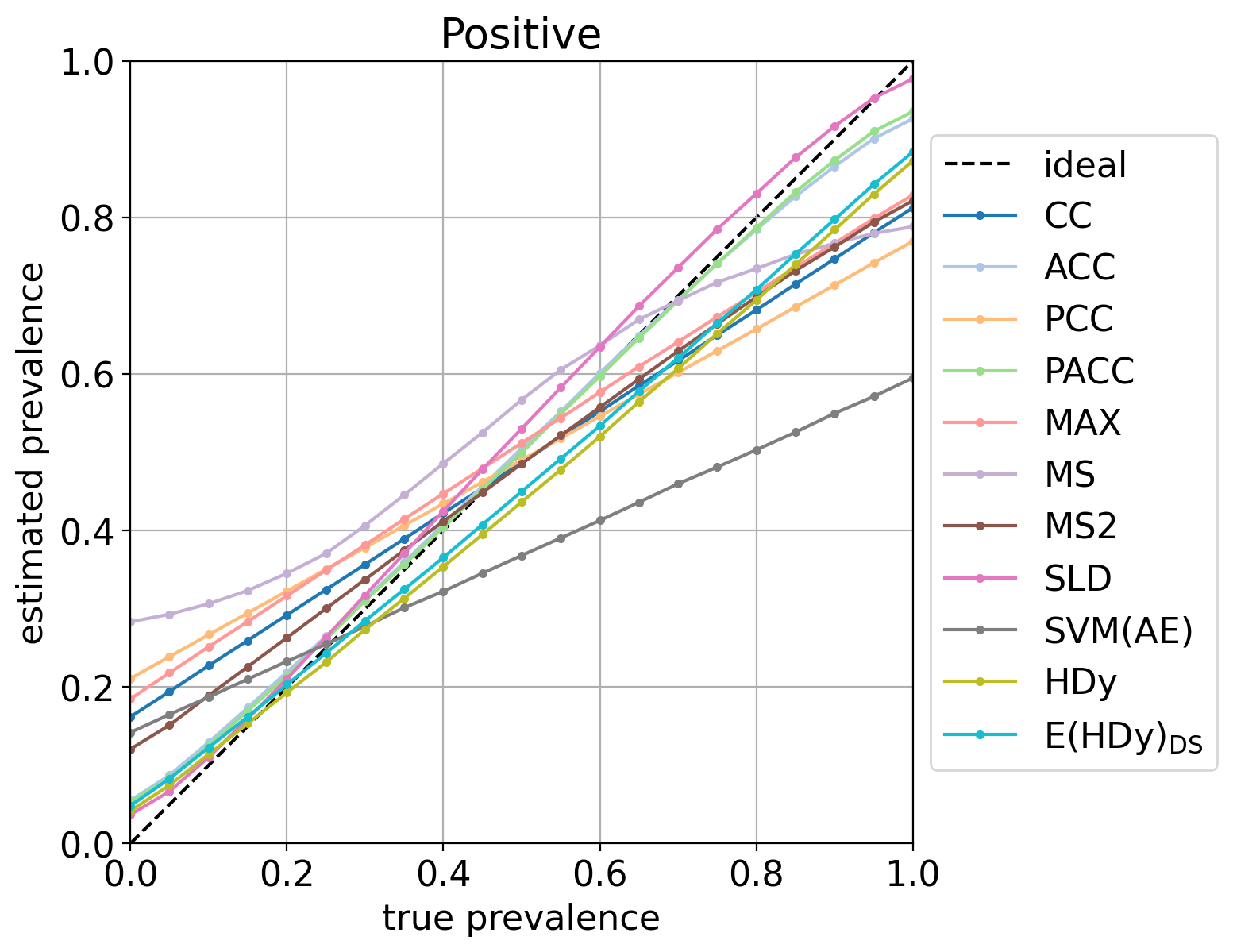}
  \includegraphics[width=0.5\textwidth]{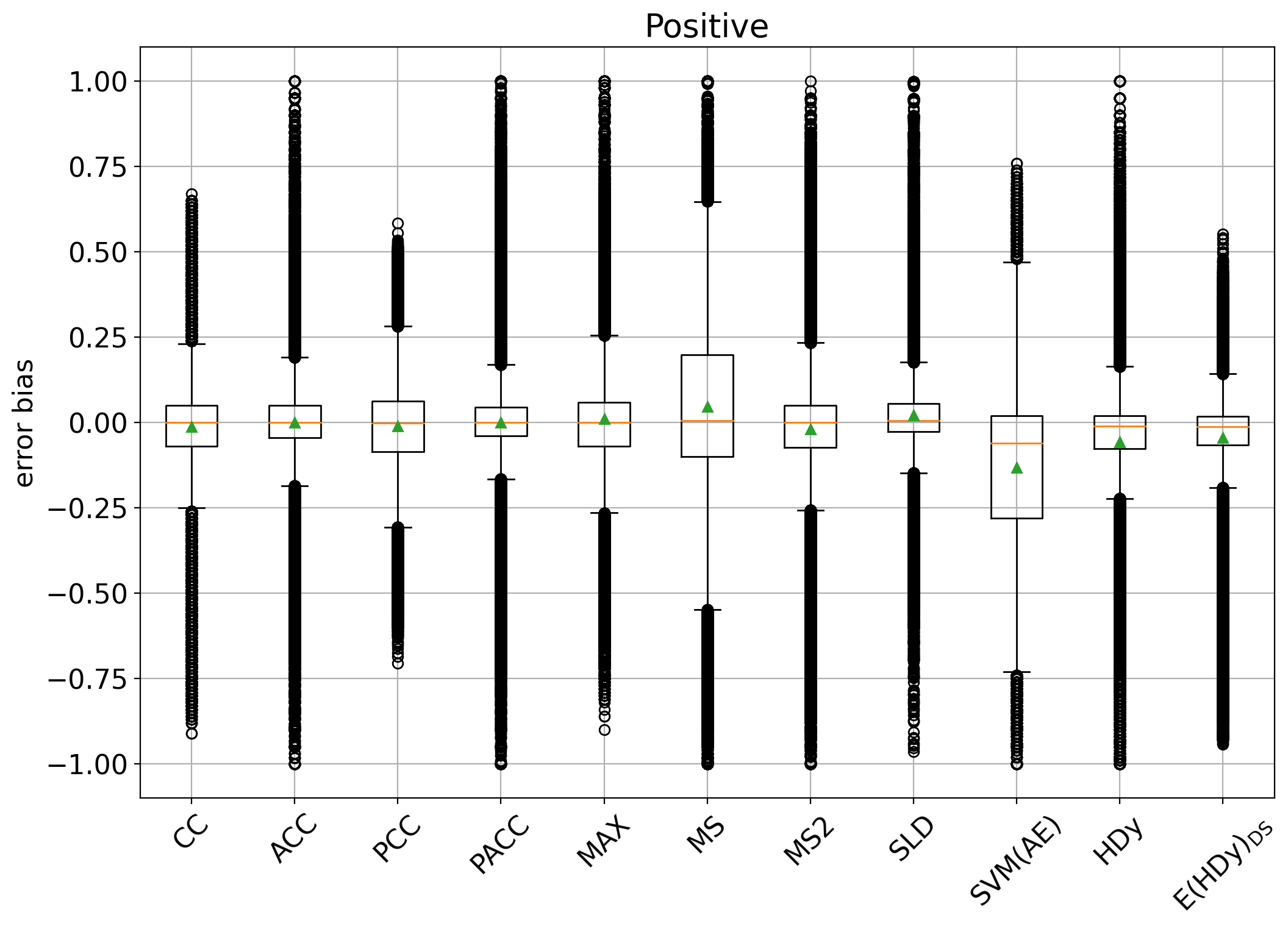}\\
  \includegraphics[width=0.45\textwidth]{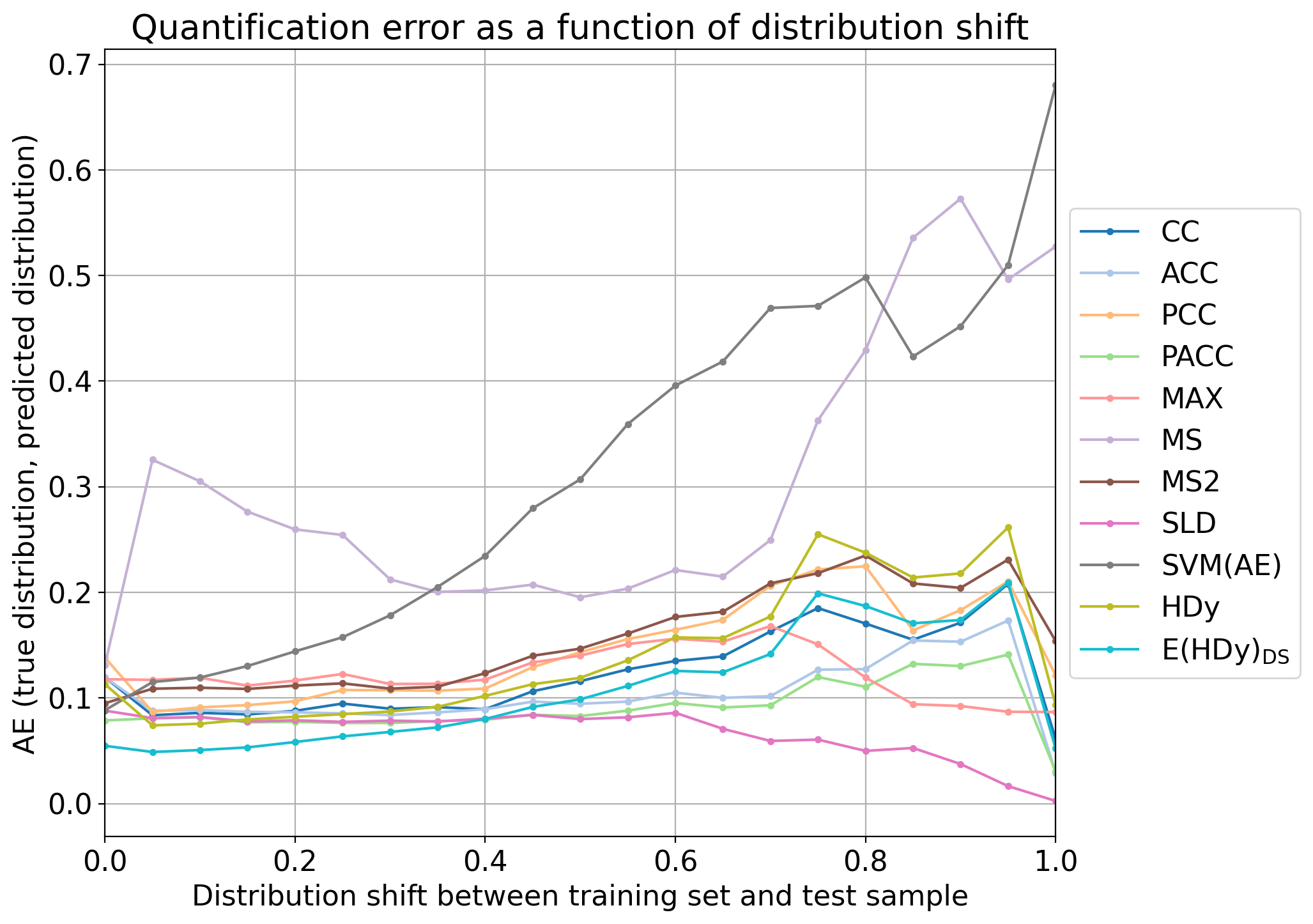}
  \includegraphics[width=0.52\textwidth]{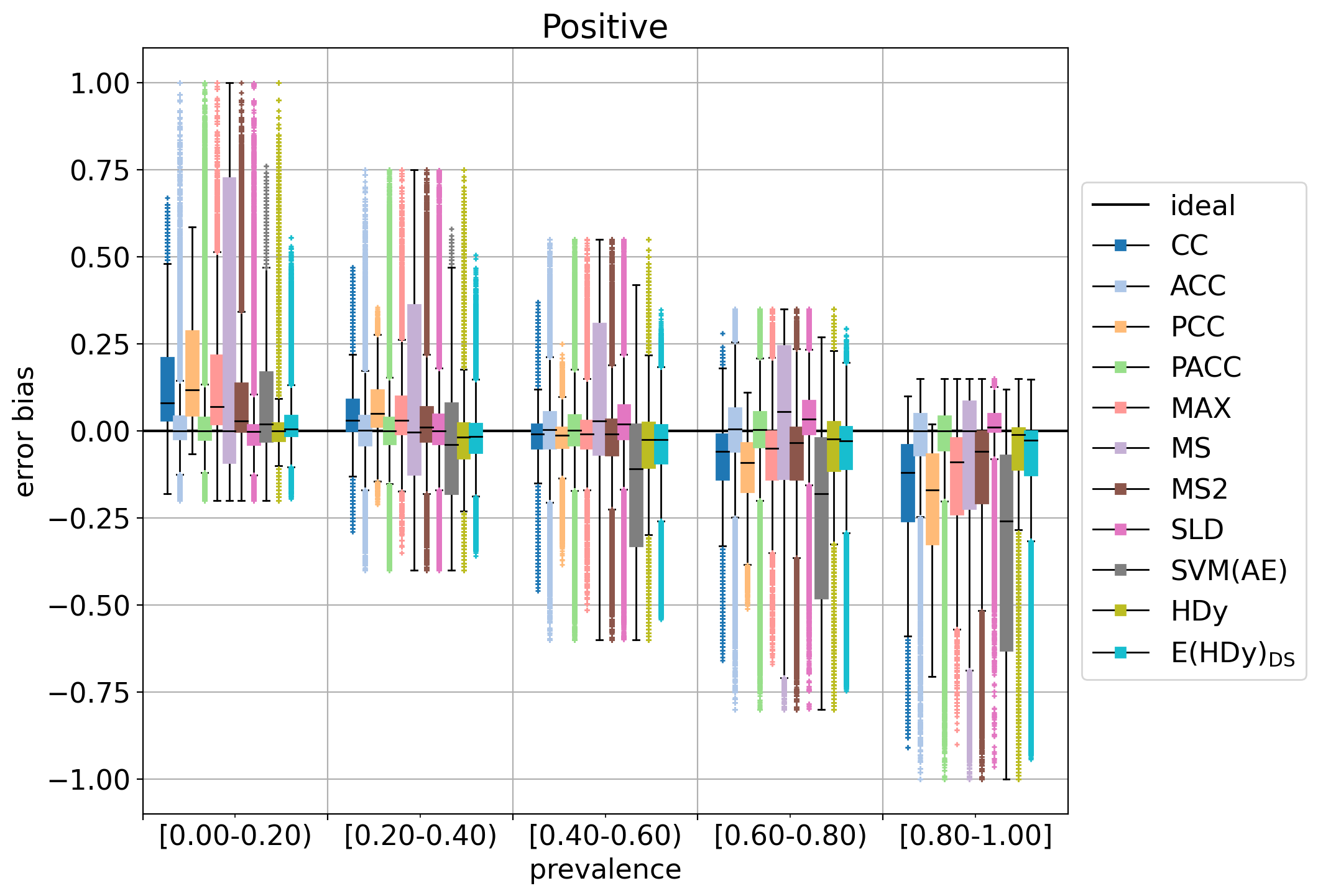}
  \caption{Examples of plots generated by QuaPy: Diagonal plot (top
  left), Error-by-Shift plot (bottom left), Global Bias-Box plot (top
  right), and Local (5 bins) Bias-Box plot (bottom right).}
  \label{fig:plots2}
\end{figure*}

\noindent QuaPy implements some plotting functions that can be useful
in displaying the performance of the tested quantification methods:

\begin{itemize}

\item\textbf{Diagonal plot}: The diagonal plot shows a very insightful
  view of the quantifier's performance, i.e., it plots the predicted
  class prevalence (on the y-axis) against the true class prevalence
  (on the x-axis), averaging across all samples characterized by the
  same true prevalence. Unfortunately, this visualization device is
  inherently limited to binary quantification (one can simply generate
  as many diagonal plots as there are classes, though, by indicating
  which class should be considered the target of the
  plot). %(Figure~\ref{fig:plots}, left).
    
\item\textbf{Error-by-Shift plot}: This plot displays the
  quantification error made by a quantifier as a function of the
  distribution shift between the training set and the test sample,
  averaging across all samples characterized by the same amount of
  distribution shift. Both quantification error and distribution shift
  can be measured in terms of any measure among those described in
  Section~\ref{sec:eval}, and can be computed and plotted both in the
  binary case and in the non-binary
  case. % (Figure~\ref{fig:plots}, center).
    
\item\textbf{Bias-Box plot}: This plot aims at displaying, by means of
  box plots, the bias that any quantifier exhibits with respect to the
  training class prevalence values. %(Figure~\ref{fig:plots}, right).
  The bias can be broken down into different bins, e.g.,
  distinguishing the bias in cases of low, medium, and high prevalence
  shift.
\end{itemize}
\noindent In Figure~\ref{fig:plots2} we show examples of each of the
above types of plot, as resulting from the experiments that we will
discuss in Section~\ref{sec:experiments}.

% \begin{figure*}
%   \centering \includegraphics[height=4cm]{Figure/bin_diag.png}
%   \includegraphics[height=4cm]{Figure/err_shift.png}
%   \includegraphics[height=4cm]{Figure/bin_bias.png}
%   \caption{Plot examples generated by QuaPy on the CC-based
%   approaches. From left to right: diagonal plot, quantification
%   error as a function of distribution shift, and box plots error
%   bias.}
%   \label{fig:plots}
% \end{figure*}

% --------------------------------------------------------------------

\section{Experiments}
\label{sec:experiments}

\noindent In this section we present some experiments that we have
carried out in order to showcase some among the features of QuaPy.
The code to replicate all these experiments, and to generate the
relative tables and plots, can be accessed via GitHub.\footnote{See
the files \texttt{uci\_experiments.py} (runs all experiments),
\texttt{uci\_tables.py} (generates Table \ref{tab:results} directly in
\LaTeX), and \texttt{uci\_plots.py} (generates all plots from Figure
\ref{fig:plots2}) included in the folder \texttt{wiki\_examples/} of
the repository \url{https://github.com/HLT-ISTI/QuaPy.wiki.git}}

As the datasets, we consider the set of UCI Machine Learning datasets
used in~\cite{Perez-Gallego:2017wt}, consisting of 30 binary datasets
(see Section~\ref{sec:datasets}).\footnote{In their
study,~\cite{Perez-Gallego:2017wt} used 32 datasets. However, we have
not been able to locate datasets ``diabetes'' and ``phoneme'' in the
UCI ML repository.} Following~\cite{Perez-Gallego:2017wt}, we remove
the ``frustratingly easy'' datasets \texttt{acute.a},
\texttt{acute.b}, and \texttt{iris.1}, where even a trivial CC
approach manages to yield zero quantification error. The datasets do
not come with a predefined train/test split; we thus carry out an
evaluation based on $5$-fold cross-validation and report the average
quantification error across the $5$ test folds. Each iteration thus
defines a training set $L$ (4 folds) and a test set $U$ (1 fold).  We
choose AE as our error metric and adopt the APP protocol for
evaluation.  For each method and test set $U$ we generate $m=100$
different random samples of $q=100$ instances each, at prevalence
values in the range $[0.00, 0.05, \ldots, 0.95, 1.00]$ via selective
undersampling, and report the resulting MAE value. Each MAE value we
report corresponds to the average of 10,500 experiments (100 samples
$\times$ 21 class prevalence values $\times$ 5 folds).

For model selection, we split the training set $L$ into a proper
training set $L_{\mathrm{Tr}}$ (consisting of 60\% of $L$) and a
held-out validation set $L_{\mathrm{Va}}$ (the remaining 40\%) in a
stratified way. For each combination of hyperparameters we train the
model using $L_{\mathrm{Tr}}$ and evaluate the performance on
$L_{\mathrm{Va}}$ in terms of MAE by following the APP
protocol~\cite{moreo2021re}; in this case we use $q=100$ and $m=25$.
Once the best values of the hyperparameters have been identified, we
re-train the method using the entire training set.

All quantifiers we consider in this demonstration are either
aggregative quantifiers or ensembles of aggregative base quantifiers,
which means that all of them rely on an underlying classifier. We
consider Logistic Regression (LR) as our default classifier-training
algorithm in all cases, except for the methods from the ``explicit
loss minimization'' camp, which instead natively rely on \svmperf.
The set of hyperparameters to optimize include the regularization
parameter $C$ (common to LR and SVMs), taking values in
$\{10^{-3}, 10^{-2}, \ldots, 10^{2}, 10^{3}\}$, and the parameter
\texttt{class\_weight} (only for LR), which may take values
\texttt{balanced} (which has the effect of giving more weight to test
examples from less frequent classes)
% \aesucomment{(thus balancing the penalty costs that
% misclassifications from each class bring to bear on the
% regularization} \fabsebcomment{Unclear})\aesucomment{Riscritta
% così?}  (weighting the cost of errors with a value that is inversely
% proportional to the number of examples belonging the correct class
% of the error, thus giving all classes equal relevance in the
% regularization process)
or \texttt{None} (which has the effect of giving the same weight to
all test examples).
% (giving any error the same weight, irrespective of the class
% involved, thus implicitly giving more relevance to more popular
% classes).

As the learning methods we consider CC, its variants PCC, ACC, PACC,
Forman's variants\footnote{To avoid clutter, we report only the three
Forman's variants that have worked best in most of the experiments
reported in~\cite{Forman:2006uf}. Additional experiments that we have
run, and that we do not report in this paper, confirm that T50 and X
perform much worse than the other methods.} MAX, MS, MS2, the
expectation-maximization-based SLD method,\footnote{Despite the fact
that classifiers trained by LR are considered inherently
well-calibrated (see, e.g.,
\url{https://scikit-learn.org/stable/modules/calibration.html}),
\cite{Molinari:2021tf} has found that re-calibrating LR brings
additional benefits to SLD. In our experiments we thus instantiate SLD
with a re-calibrated version of LR, and we indeed observe this to
improve results noticeably. However, re-calibrating does not deliver
any improvement for any other probabilistic quantifier that we test
here, and instead shows a tendency to deteriorate the results. For
this reason, we use a re-calibrated LR only for SLD, and a
``standard'' LR %elsewhere.
in all other cases.} the mixture model HDy, SVM(AE) as the
representative of the ``explicit loss minimization''
family\footnote{Among all ELM-based methods, we choose the one that
minimizes the same loss that we adopt for evaluating the results. We
do not consider other variants (SVM(Q), SVM(KLD), SVM(NKLD), SVM(RAE))
since, in recent evaluations (see, e.g.,
\cite{Moreo:2020mk,moreo2021re}), they have consistently
underperformed other competitors.}, and \ehdy\ as the representative
of ensemble methods (since it is the one which fared best in the
experiments of~\cite{Perez-Gallego:2019vl}). For \ehdy\, we set the
number of base quantifiers to \texttt{size=30} and the number of
members to be selected dynamically to \texttt{red\_size=15}, and
perform model selection independently for each base member.

% \fabsebcomment{Da ricontrollare da qui.}
Table~\ref{tab:results} reports the AE results of this
experimentation. Our results are fairly consistent with those reported
in~\cite{Moreo:2020mk,moreo2021re}, and seem to indicate that the
strongest method of all is SLD, which obtains the best average MAE
result, the best average rank, and is the best method on 13 datasets
out of 30.  Methods \ehdy\ (8 times best method), PACC (4 times best
method), and (to a lesser extent) ACC (2 times best method), also seem
to perform very well, obtaining average ranks not statistically
significantly different from the best average rank (obtained by SLD).
% In line with~\cite{moreo2021re}, CC and its three main variants (CC,
% ACC, PCC, PACC) perform reasonably well when properly optimized
% (i.e., when model selection is carried out for quantification, and
% not for classification).
Method
% s TS50, TSX, and
SVM(AE) tends to produce results that are markedly worse than the rest
of competitors. In line with the observations of~\cite{Schumacher},
none of the variants MAX,
% TSX, TS50,
MS, MS2 manages to improve over ACC.
% \fabsebcomment{T50 l'abbiamo levato?}
Also in line with the findings of \cite{Perez-Gallego:2019vl}, the
ensemble \ehdy\ clearly outperforms the base quantifier HDy it is
built upon.

\begin{table*}[ht!]
  \centering 
                  \resizebox{\textwidth}{!}{%
                        \begin{tabular}{|c||c|c|c|c|c|c|c|c|c|c|c||} \hline
                          & \multicolumn{11}{c||}{Quantification methods} \\ \hline
                   & \side{CC} & \side{ACC} & \side{PCC} & \side{PACC} & \side{MAX} & \side{MS} & \side{MS2} & \side{SLD} & \side{SVM(AE)} & \side{HDy} & \side{E(HDy)$_\mathrm{DS}$} \\\hline
BALANCE.1 &  0.039$\pdag$ \cellcolor{red!14} &  0.032$\pdag$ \cellcolor{green!8} &  0.049$\pdag$ \cellcolor{red!50} &  0.037$\pdag$ \cellcolor{red!8} &  0.040$\pdag$ \cellcolor{red!19} &  0.046$\pdag$ \cellcolor{red!39} &  0.036$\pdag$ \cellcolor{red!6} &  0.025$\pdag$ \cellcolor{green!31} &  0.035$\pdag$ \cellcolor{red!4} &  0.022$\pdag$ \cellcolor{green!41} & \textbf{0.020}$\pdag$ \cellcolor{green!50} \\\hline
BALANCE.2 &  0.314$\pdag$ \cellcolor{green!29} &  0.379$\pdag$ \cellcolor{green!1} & \textbf{0.264}$\pdag$ \cellcolor{green!50} &  0.432$\pdag$ \cellcolor{red!21} &  0.465$\pdag$ \cellcolor{red!35} &  0.288$\pdag$ \cellcolor{green!39} &  0.331$\pdag$ \cellcolor{green!21} &  0.372$\pdag$ \cellcolor{green!4} &  0.500$\pdag$ \cellcolor{red!50} &  0.470$\pdag$ \cellcolor{red!37} &  0.355$\pdag$ \cellcolor{green!11} \\\hline
BALANCE.3 &  0.039$\pdag$ \cellcolor{green!0} &  0.020$\pdag$ \cellcolor{green!39} &  0.045$\pdag$ \cellcolor{red!12} &  0.021$\pdag$ \cellcolor{green!37} &  0.040$\pdag$ \cellcolor{red!1} &  0.046$\pdag$ \cellcolor{red!13} &  0.036$\pdag$ \cellcolor{green!6} &  0.018$\pdag$ \cellcolor{green!42} &  0.064$\pdag$ \cellcolor{red!50} &  0.017$\pdag$ \cellcolor{green!43} & \textbf{0.014}$\pdag$ \cellcolor{green!50} \\\hline
BREAST-CANCER &  0.022$\pdag$ \cellcolor{green!48} &  0.025$\pdag$ \cellcolor{green!46} &  0.029$\pdag$ \cellcolor{green!42} &  0.023$\pdag$ \cellcolor{green!47} &  0.028$\pdag$ \cellcolor{green!43} &  0.021$\pdag$ \cellcolor{green!49} &  0.023$\pdag$ \cellcolor{green!47} & \textbf{0.020}$\pdag$ \cellcolor{green!50} &  0.144$\pdag$ \cellcolor{red!50} &  0.029$\pdag$ \cellcolor{green!42} &  0.026$\pdag$ \cellcolor{green!44} \\\hline
CMC.1 &  0.194$\pdag$ \cellcolor{red!25} &  0.108$\pdag$ \cellcolor{green!39} &  0.226$\pdag$ \cellcolor{red!48} &  0.117$\pdag$ \cellcolor{green!32} &  0.191$\pdag$ \cellcolor{red!22} &  0.195$\pdag$ \cellcolor{red!25} &  0.178$\pdag$ \cellcolor{red!13} & \textbf{0.094}$\pdag$ \cellcolor{green!50} &  0.227$\pdag$ \cellcolor{red!50} &  0.156$\pdag$ \cellcolor{green!3} &  0.126$\pdag$ \cellcolor{green!25} \\\hline
CMC.2 &  0.178$\pdag$ \cellcolor{green!29} &  0.138$\pdag$ \cellcolor{green!40} &  0.220$\pdag$ \cellcolor{green!19} & \textbf{0.098}$\pdag$ \cellcolor{green!50} &  0.271$\pdag$ \cellcolor{green!6} &  0.500$\pdag$ \cellcolor{red!50} &  0.427$\pdag$ \cellcolor{red!31} &  0.105$\pdag$ \cellcolor{green!48} &  0.449$\pdag$ \cellcolor{red!37} &  0.118$\pdag$ \cellcolor{green!44} &  0.103$\pdag$ \cellcolor{green!48} \\\hline
CMC.3 &  0.211$\pdag$ \cellcolor{green!14} &  0.172$\pdag$ \cellcolor{green!30} &  0.239$\pdag$ \cellcolor{green!3} &  0.127$\pdag$ \cellcolor{green!48} &  0.254$\pdag$ \cellcolor{red!2} &  0.376$\pdag$ \cellcolor{red!50} &  0.353$\pdag$ \cellcolor{red!41} &  0.124$^{\ddag}$ \cellcolor{green!49} &  0.336$\pdag$ \cellcolor{red!34} &  0.136$\pdag$ \cellcolor{green!44} & \textbf{0.122}$\pdag$ \cellcolor{green!50} \\\hline
CTG.1 &  0.037$\pdag$ \cellcolor{green!24} &  0.020$\pdag$ \cellcolor{green!46} &  0.050$\pdag$ \cellcolor{green!7} &  0.020$\pdag$ \cellcolor{green!46} &  0.041$\pdag$ \cellcolor{green!18} &  0.033$\pdag$ \cellcolor{green!29} &  0.035$\pdag$ \cellcolor{green!26} & \textbf{0.017}$\pdag$ \cellcolor{green!50} &  0.094$\pdag$ \cellcolor{red!50} &  0.028$\pdag$ \cellcolor{green!36} &  0.018$\pdag$ \cellcolor{green!48} \\\hline
CTG.2 &  0.048$\pdag$ \cellcolor{green!47} &  0.040$\pdag$ \cellcolor{green!48} &  0.078$\pdag$ \cellcolor{green!42} &  0.045$\pdag$ \cellcolor{green!47} &  0.048$\pdag$ \cellcolor{green!47} &  0.653$\pdag$ \cellcolor{red!50} &  0.059$\pdag$ \cellcolor{green!45} & \textbf{0.030}$\pdag$ \cellcolor{green!50} &  0.152$\pdag$ \cellcolor{green!30} &  0.045$\pdag$ \cellcolor{green!47} &  0.040$\pdag$ \cellcolor{green!48} \\\hline
CTG.3 &  0.047$\pdag$ \cellcolor{green!46} &  0.044$\pdag$ \cellcolor{green!46} &  0.050$\pdag$ \cellcolor{green!45} &  0.043$\pdag$ \cellcolor{green!46} &  0.045$\pdag$ \cellcolor{green!46} &  0.649$\pdag$ \cellcolor{red!50} &  0.061$\pdag$ \cellcolor{green!43} & \textbf{0.022}$\pdag$ \cellcolor{green!50} &  0.113$\pdag$ \cellcolor{green!35} &  0.053$\pdag$ \cellcolor{green!45} &  0.045$\pdag$ \cellcolor{green!46} \\\hline
GERMAN &  0.151$\pdag$ \cellcolor{green!15} &  0.142$\pdag$ \cellcolor{green!20} &  0.191$\pdag$ \cellcolor{red!7} & \textbf{0.092}$\pdag$ \cellcolor{green!50} &  0.154$\pdag$ \cellcolor{green!13} &  0.125$\pdag$ \cellcolor{green!30} &  0.134$\pdag$ \cellcolor{green!25} &  0.101$\pdag$ \cellcolor{green!44} &  0.262$\pdag$ \cellcolor{red!50} &  0.165$\pdag$ \cellcolor{green!7} &  0.113$\pdag$ \cellcolor{green!37} \\\hline
HABERMAN &  0.231$\pdag$ \cellcolor{green!39} &  0.190$^{\ddag}$ \cellcolor{green!49} &  0.237$\pdag$ \cellcolor{green!37} &  0.267$\pdag$ \cellcolor{green!29} &  0.242$\pdag$ \cellcolor{green!36} &  0.572$\pdag$ \cellcolor{red!50} &  0.244$\pdag$ \cellcolor{green!35} & \textbf{0.190}$\pdag$ \cellcolor{green!50} &  0.283$\pdag$ \cellcolor{green!25} &  0.399$\pdag$ \cellcolor{red!4} &  0.324$\pdag$ \cellcolor{green!14} \\\hline
IONOSPHERE &  0.111$\pdag$ \cellcolor{green!29} & \textbf{0.074}$\pdag$ \cellcolor{green!50} &  0.116$\pdag$ \cellcolor{green!26} &  0.084$\pdag$ \cellcolor{green!44} &  0.124$\pdag$ \cellcolor{green!22} &  0.209$\pdag$ \cellcolor{red!23} &  0.089$\pdag$ \cellcolor{green!41} &  0.075$^{\ddag}$ \cellcolor{green!49} &  0.256$\pdag$ \cellcolor{red!50} &  0.104$\pdag$ \cellcolor{green!33} &  0.082$\pdag$ \cellcolor{green!45} \\\hline
IRIS.2 &  0.201$\pdag$ \cellcolor{green!14} &  0.241$\pdag$ \cellcolor{green!4} &  0.195$\pdag$ \cellcolor{green!15} &  0.183$\pdag$ \cellcolor{green!18} &  0.251$\pdag$ \cellcolor{green!1} &  0.412$\pdag$ \cellcolor{red!37} &  0.256$\pdag$ \cellcolor{green!0} &  0.215$\pdag$ \cellcolor{green!10} &  0.461$\pdag$ \cellcolor{red!50} &  0.075$\pdag$ \cellcolor{green!45} & \textbf{0.056}$\pdag$ \cellcolor{green!50} \\\hline
IRIS.3 & \textbf{0.019}$\pdag$ \cellcolor{green!50} &  0.074$\pdag$ \cellcolor{green!20} &  0.044$\pdag$ \cellcolor{green!36} &  0.071$\pdag$ \cellcolor{green!22} &  0.054$\pdag$ \cellcolor{green!31} &  0.134$\pdag$ \cellcolor{red!11} &  0.024$\pdag$ \cellcolor{green!47} &  0.057$\pdag$ \cellcolor{green!29} &  0.205$\pdag$ \cellcolor{red!50} &  0.069$\pdag$ \cellcolor{green!23} &  0.047$\pdag$ \cellcolor{green!35} \\\hline
MAMMOGRAPHIC &  0.090$\pdag$ \cellcolor{red!7} &  0.048$\pdag$ \cellcolor{green!33} &  0.130$\pdag$ \cellcolor{red!46} &  0.040$\pdag$ \cellcolor{green!41} &  0.091$\pdag$ \cellcolor{red!8} &  0.059$\pdag$ \cellcolor{green!23} &  0.060$\pdag$ \cellcolor{green!22} &  0.036$\pdag$ \cellcolor{green!45} &  0.134$\pdag$ \cellcolor{red!50} &  0.044$\pdag$ \cellcolor{green!37} & \textbf{0.031}$\pdag$ \cellcolor{green!50} \\\hline
PAGEBLOCKS.5 &  0.048$\pdag$ \cellcolor{green!48} & \textbf{0.040}$\pdag$ \cellcolor{green!50} &  0.067$\pdag$ \cellcolor{green!43} &  0.041$^{\ddag}$ \cellcolor{green!49} &  0.066$\pdag$ \cellcolor{green!44} &  0.474$\pdag$ \cellcolor{red!50} &  0.115$\pdag$ \cellcolor{green!32} &  0.070$\pdag$ \cellcolor{green!43} &  0.342$\pdag$ \cellcolor{red!19} &  0.085$\pdag$ \cellcolor{green!39} &  0.066$\pdag$ \cellcolor{green!44} \\\hline
SEMEION &  0.042$\pdag$ \cellcolor{green!47} &  0.049$\pdag$ \cellcolor{green!45} &  0.058$\pdag$ \cellcolor{green!44} &  0.040$\pdag$ \cellcolor{green!47} &  0.038$\pdag$ \cellcolor{green!48} &  0.500$\pdag$ \cellcolor{red!50} &  0.074$\pdag$ \cellcolor{green!40} & \textbf{0.030}$\pdag$ \cellcolor{green!50} &  0.070$\pdag$ \cellcolor{green!41} &  0.037$\pdag$ \cellcolor{green!48} &  0.047$\pdag$ \cellcolor{green!46} \\\hline
SONAR &  0.135$\pdag$ \cellcolor{green!40} &  0.200$\pdag$ \cellcolor{green!12} &  0.163$\pdag$ \cellcolor{green!28} &  0.119$\pdag$ \cellcolor{green!47} &  0.145$\pdag$ \cellcolor{green!36} &  0.171$\pdag$ \cellcolor{green!25} &  0.159$\pdag$ \cellcolor{green!30} & \textbf{0.114}$\pdag$ \cellcolor{green!50} &  0.346$\pdag$ \cellcolor{red!50} &  0.136$\pdag$ \cellcolor{green!40} &  0.131$\pdag$ \cellcolor{green!42} \\\hline
SPAMBASE &  0.042$\pdag$ \cellcolor{green!38} &  0.026$\pdag$ \cellcolor{green!47} &  0.066$\pdag$ \cellcolor{green!24} & \textbf{0.022}$\pdag$ \cellcolor{green!50} &  0.049$\pdag$ \cellcolor{green!34} &  0.070$\pdag$ \cellcolor{green!22} &  0.037$\pdag$ \cellcolor{green!41} &  0.031$\pdag$ \cellcolor{green!44} &  0.196$\pdag$ \cellcolor{red!50} &  0.025$\pdag$ \cellcolor{green!48} &  0.024$\pdag$ \cellcolor{green!48} \\\hline
SPECTF &  0.143$\pdag$ \cellcolor{green!42} &  0.155$\pdag$ \cellcolor{green!40} &  0.178$\pdag$ \cellcolor{green!35} &  0.133$\pdag$ \cellcolor{green!44} &  0.276$\pdag$ \cellcolor{green!16} &  0.620$\pdag$ \cellcolor{red!50} &  0.182$\pdag$ \cellcolor{green!35} & \textbf{0.105}$\pdag$ \cellcolor{green!50} &  0.296$\pdag$ \cellcolor{green!13} &  0.420$\pdag$ \cellcolor{red!11} &  0.231$\pdag$ \cellcolor{green!25} \\\hline
TICTACTOE &  0.024$\pdag$ \cellcolor{green!48} &  0.019$\pdag$ \cellcolor{green!49} &  0.024$\pdag$ \cellcolor{green!48} & \textbf{0.014}$\pdag$ \cellcolor{green!50} &  0.024$\pdag$ \cellcolor{green!48} &  0.136$\pdag$ \cellcolor{green!24} &  0.024$\pdag$ \cellcolor{green!48} &  0.019$\pdag$ \cellcolor{green!48} &  0.500$\pdag$ \cellcolor{red!50} &  0.018$\pdag$ \cellcolor{green!49} &  0.019$\pdag$ \cellcolor{green!49} \\\hline
TRANSFUSION &  0.178$\pdag$ \cellcolor{green!28} &  0.139$\pdag$ \cellcolor{green!37} &  0.215$\pdag$ \cellcolor{green!19} &  0.097$\pdag$ \cellcolor{green!47} &  0.220$\pdag$ \cellcolor{green!18} &  0.510$\pdag$ \cellcolor{red!50} &  0.433$\pdag$ \cellcolor{red!31} & \textbf{0.087}$\pdag$ \cellcolor{green!50} &  0.442$\pdag$ \cellcolor{red!33} &  0.246$\pdag$ \cellcolor{green!12} &  0.166$\pdag$ \cellcolor{green!31} \\\hline
WDBC &  0.034$\pdag$ \cellcolor{green!26} &  0.036$\pdag$ \cellcolor{green!24} &  0.034$\pdag$ \cellcolor{green!26} &  0.027$\pdag$ \cellcolor{green!34} &  0.038$\pdag$ \cellcolor{green!21} &  0.096$\pdag$ \cellcolor{red!50} &  0.029$\pdag$ \cellcolor{green!33} &  0.025$\pdag$ \cellcolor{green!37} &  0.056$\pdag$ \cellcolor{red!0} &  0.019$\pdag$ \cellcolor{green!45} & \textbf{0.015}$\pdag$ \cellcolor{green!50} \\\hline
WINE.1 &  0.029$\pdag$ \cellcolor{green!40} &  0.025$\pdag$ \cellcolor{green!44} &  0.025$\pdag$ \cellcolor{green!44} &  0.030$\pdag$ \cellcolor{green!40} &  0.033$\pdag$ \cellcolor{green!37} &  0.133$\pdag$ \cellcolor{red!50} &  0.030$\pdag$ \cellcolor{green!40} &  0.044$\pdag$ \cellcolor{green!28} &  0.062$\pdag$ \cellcolor{green!12} &  0.040$\pdag$ \cellcolor{green!31} & \textbf{0.019}$\pdag$ \cellcolor{green!50} \\\hline
WINE.2 &  0.026$\pdag$ \cellcolor{green!43} &  0.048$\pdag$ \cellcolor{green!11} &  0.043$\pdag$ \cellcolor{green!18} &  0.052$\pdag$ \cellcolor{green!4} &  0.045$\pdag$ \cellcolor{green!15} &  0.088$\pdag$ \cellcolor{red!50} &  0.041$\pdag$ \cellcolor{green!21} &  0.046$\pdag$ \cellcolor{green!13} &  0.051$\pdag$ \cellcolor{green!5} &  0.032$\pdag$ \cellcolor{green!35} & \textbf{0.022}$\pdag$ \cellcolor{green!50} \\\hline
WINE.3 &  0.031$\pdag$ \cellcolor{green!41} &  0.040$\pdag$ \cellcolor{green!36} & \textbf{0.016}$\pdag$ \cellcolor{green!50} &  0.033$\pdag$ \cellcolor{green!40} &  0.028$\pdag$ \cellcolor{green!43} &  0.190$\pdag$ \cellcolor{red!50} &  0.029$\pdag$ \cellcolor{green!42} &  0.061$\pdag$ \cellcolor{green!24} &  0.018$^{\dag}$ \cellcolor{green!49} &  0.018$\pdag$ \cellcolor{green!49} &  0.025$\pdag$ \cellcolor{green!45} \\\hline
WINE-Q-RED &  0.140$\pdag$ \cellcolor{red!0} &  0.076$\pdag$ \cellcolor{green!37} &  0.183$\pdag$ \cellcolor{red!26} &  0.059$\pdag$ \cellcolor{green!48} &  0.141$\pdag$ \cellcolor{red!1} &  0.065$\pdag$ \cellcolor{green!44} &  0.099$\pdag$ \cellcolor{green!24} & \textbf{0.056}$\pdag$ \cellcolor{green!50} &  0.222$\pdag$ \cellcolor{red!50} &  0.065$\pdag$ \cellcolor{green!44} &  0.058$\pdag$ \cellcolor{green!48} \\\hline
WINE-Q-WHITE &  0.150$\pdag$ \cellcolor{green!1} &  0.077$\pdag$ \cellcolor{green!40} &  0.194$\pdag$ \cellcolor{red!21} &  0.064$\pdag$ \cellcolor{green!47} &  0.149$\pdag$ \cellcolor{green!1} &  0.113$\pdag$ \cellcolor{green!21} &  0.124$\pdag$ \cellcolor{green!15} & \textbf{0.059}$\pdag$ \cellcolor{green!50} &  0.247$\pdag$ \cellcolor{red!50} &  0.072$\pdag$ \cellcolor{green!42} &  0.066$\pdag$ \cellcolor{green!46} \\\hline
YEAST &  0.155$\pdag$ \cellcolor{green!21} &  0.107$\pdag$ \cellcolor{green!36} &  0.197$\pdag$ \cellcolor{green!7} &  0.071$\pdag$ \cellcolor{green!48} &  0.159$\pdag$ \cellcolor{green!19} &  0.233$\pdag$ \cellcolor{red!3} &  0.235$\pdag$ \cellcolor{red!4} & \textbf{0.066}$\pdag$ \cellcolor{green!50} &  0.378$\pdag$ \cellcolor{red!50} &  0.073$\pdag$ \cellcolor{green!47} &  0.071$\pdag$ \cellcolor{green!48} \\\hline
\hline
Average &  0.104$^{\ddag}$ \cellcolor{green!35} &  0.093$^{\ddag}$ \cellcolor{green!41} &  0.121$^{\dag}$ \cellcolor{green!25} &  0.083$^{\ddag}$ \cellcolor{green!46} &  0.125$^{\dag}$ \cellcolor{green!23} &  0.257$\pdag$ \cellcolor{red!50} &  0.132$^{\dag}$ \cellcolor{green!19} & \textbf{0.077}$\pdag$ \cellcolor{green!50} &  0.231$\pdag$ \cellcolor{red!35} &  0.107$^{\ddag}$ \cellcolor{green!33} &  0.083$^{\ddag}$ \cellcolor{green!46} \\\hline
Rank Average &  5.733$\pdag$ \cellcolor{green!11} &  4.967$^{\dag}$ \cellcolor{green!22} &  7.033$\pdag$ \cellcolor{red!7} &  3.900$^{\ddag}$ \cellcolor{green!38} &  7.133$\pdag$ \cellcolor{red!9} &  9.033$\pdag$ \cellcolor{red!37} &  6.733$\pdag$ \cellcolor{red!3} & \textbf{3.133}$\pdag$ \cellcolor{green!50} &  9.900$\pdag$ \cellcolor{red!50} &  5.233$\pdag$ \cellcolor{green!18} &  3.200$^{\ddag}$ \cellcolor{green!49} \\\hline

            \end{tabular}%
            }
  \caption{Values of AE obtained in our experiments; each value is the
  average across 10,500 values, each obtained on a different sample.
  \textbf{Boldface} indicates the best method for a given dataset.
  Superscripts $\dag$ and $\ddag$ denote the methods (if any) whose
  scores are \emph{not} statistically significantly different from the
  best one according to a paired sample, two-tailed t-test at
  different confidence levels: symbol $\dag$ indicates $0.001<p$-value
  $<0.05$ while symbol $\ddag$ indicates $0.05\leq p$-value. The
  absence of any such symbol indicates $p$-value $\leq 0.001$ (i.e.,
  that the performance of the method is statistically significantly
  different from that of the best method). For ease of readability,
  for each dataset we colour-code cells via intense green for the best
  result, intense red for the worst result, and an interpolated tone
  for the scores in-between.}
  \label{tab:results}
\end{table*}

Figure~\ref{fig:plots2} shows examples of plots (3 out of 4 plots are
only for the \textsf{Positive} class) generated using QuaPy.
The Diagonal plot (the results are averages across all samples
characterized by the same true class prevalence values) reveals that,
for high prevalence values of the \textsf{Positive} class, SLD tends
to slightly overestimate these class prevalence values while most
other methods tend instead to underestimate them. For low prevalence
values of the \textsf{Positive} class, methods MAX, MS, MS2, PCC, and
CC tend to overestimate these prevalence values.
The Error-by-Shift plot (bottom left) displays AE as a function of the
distribution shift between the training set and each of the test
samples. (The results are averages across all samples characterized by
the same value of distribution shift.) Here one can appreciate that
\ehdy\ excels at situations characterized by low distribution shift,
while SLD seems the most robust in dealing with high-shift scenarios.
The Bias-Box plots (top right) show the distribution of error bias
(i.e., of the signed error between the estimated prevalence value and
the true prevalence value) for all methods, as averaged across all
datasets and test samples. This diagram reveals that PACC, SLD, and
\ehdy\ are the methods displaying the lowest bias overall, given that
their boxes (delimiting the first and third quartiles) are the most
squashed, and given that their whiskers (maximum and minimum,
disregarding outliers) are the shortest. One interesting fact that is
clearly revealed by this box-plot is, in line with what reported in
\cite{Perez-Gallego:2019vl}, the ability of the ensemble method
(\ehdy) to reduce the variance of the base quantifiers it is built
upon (HDy).  It is also interesting to note how the heuristic
implemented in MS2 drastically reduces the variance produced by MS.
The last plot (bottom right) displays error bias trends with samples
binned according to their true prevalence; it clearly shows how the
``unadjusted'' methods (e.g., CC, PCC) display positive bias for low
prevalence values (thus overestimating the true prevalence) and
negative bias for high prevalence values (thus underestimating the
true prevalence), while the ``adjusted'' versions (ACC and PACC)
reduce this effect, since they tend to display box-plots centred at
zero bias in those cases.  This plot also clearly explains that MS
tends to display a huge positive bias in the low-prevalence regime,
while SVM(AE) displays a huge negative bias in the high-prevalence
regime.
% \fabsebcomment{Da ricontrollare fin qui.}

% \begin{figure}
%   \centering
%   \includegraphics[width=0.85\columnwidth]{Figures/diag_ae_pos.png}
%   \includegraphics[width=0.85\columnwidth]{Figures/error_by_drift_ae.png}
%   \includegraphics[width=0.85\columnwidth]{Figures/globalbias_ae_pos.png}
%   \includegraphics[width=0.85\columnwidth]{Figures/localbias_ae_pos.png}
%   \caption{Examples of plots generated by QuaPy. From upper to
%   bottom: Diagonal plot, Error-by-Shift plot, Bias-Box Plot
%   (global), and Bias-Box plot (local to 5 bins).}
%   \label{fig:plots2}
% \end{figure}
% \aesucomment{Ho cambiato la distribuzione delle immagini perché
% erano troppo piccole.}

Note that the results presented here are just for the purposes of
illustrating the functionality of QuaPy, and should not be taken as an
absolute statement on the relative merits of the different
quantification methods. For instance, a different batch of experiments
(those reported in~\cite{Moreo:2020mk}, and dealing with sentiment
quantification on datasets of tweets), tell a slightly different
story, since they report a much larger difference in accuracy between
top-performing methods (SLD, PACC, ACC) and lesser performing ones
(CC, PCC, SVM(AE), and others). One of the main differences between
the experiments in this paper and those in~\cite{Moreo:2020mk} is that
we here work on binary quantification only, while~\cite{Moreo:2020mk}
tackled single-label multiclass quantification (since all datasets
used there were ternary).  As always, a complete understanding of the
relative merits of different learning methods can only be obtained
through multiple, varied sets of experiments (see also
\cite{Schumacher}).

% --------------------------------------------------------------------

\section{Conclusions}
\label{sec:conclusion}

\noindent Quantification is a research topic of growing interest in
the areas of machine learning, data mining, and information retrieval.
We have presented QuaPy, an open-source, Python-based package that
makes available a rich set of quantification methods, tools,
experimental protocols, and datasets, with the goal of supporting an
efficient and scientifically correct experimentation of quantification
methods.  We think that QuaPy will be of help to machine learning
researchers that work on developing new quantification algorithms, as
it provides them with many baselines to compare against, datasets to
test their methods on, and tools that implement all the typical steps
of quantification-based experimentation, from data preparation to the
visualization of results.  We think that QuaPy will be of help also to
researchers and practitioners in other disciplines who simply need to
apply quantification in their own work, as it provides them with a
streamlined workflow, a wide choice of different approaches, and quick
access to the package thanks to the support of installation based on
\texttt{pip}.  QuaPy is an open-source project, licensed under the
BSD-3 licence; its repository will be updated following the advances
in quantification research, and it is open to contributions of new
methods, tools, and datasets.

% --------------------------------------------------------------------

% \balance
\clearpage
\newpage

\bibliographystyle{acm} \bibliography{Fabrizio,References}

% --------------------------------------------------------------------

\end{document}